\pdfoutput=1

\documentclass[11pt]{article}

\usepackage[preprint]{acl}

\usepackage{times}
\usepackage{latexsym}

\usepackage[T1]{fontenc}

\usepackage[utf8]{inputenc}

\usepackage{url}            
\usepackage{booktabs}       
\usepackage{nicefrac}       
\usepackage{microtype}      
\usepackage[table]{xcolor}  
\usepackage{amsfonts}       
\usepackage{amsmath}
\usepackage{mathtools}      
\usepackage{bbm}            
\usepackage{amssymb}        
\usepackage{graphicx}       
\usepackage{multirow}       
\usepackage{siunitx}        
\usepackage{enumitem}       
\usepackage{fontawesome5}
\usepackage{caption}        
\usepackage{float}
\usepackage{adjustbox}
\usepackage{wrapfig}
\usepackage{array}
\usepackage{makecell}
\usepackage{tikz}
\usepackage{xparse}
\usepackage{inconsolata}    
\usepackage[most]{tcolorbox}
\usepackage{xspace}
\usepackage{listings}
\usepackage{algorithm}
\usepackage{algpseudocode}
\usepackage{framed}
\usepackage{import}
\usepackage{xifthen}
\usepackage{pdfpages}
\usepackage{transparent}
\usetikzlibrary{arrows.meta, positioning, shapes.geometric, calc, patterns, shapes.callouts}

\NewDocumentCommand{\standout}{m}{\texorpdfstring{\texttt{#1}}{#1}}
\input{macros.tex}

\definecolor{benchmarkrow}{RGB}{220,230,242} 
\definecolor{userstudyrow}{RGB}{220,220,220}

\title{Thinking While Speaking: Inference-Time Knowledge Transfer for Responsive and Intelligent Conversational Voice Agents}

\author{Vidya Srinivas\thanks{Equal contribution.},  Zachary Englhardt\footnotemark[1], Vikram Iyer, Shwetak Patel \\
Paul G. Allen School of Computer Science \& Engineering\\
\texttt{\{vysri,zacharye\}@cs.washington.edu}
}

\begin{document}
\maketitle

\begin{abstract}
Voice agents face a fundamental tension: the reasoning, retrieval, and tool use that make foundation models capable are iterative and slow, while conversational interaction demands responses on a millisecond timescale. Smaller, real-time models meet the latency bar but cannot match foundation models on complex tasks, leaving current voice agents to trade away either responsiveness or capability. We introduce \textit{conversational infill}, where a small \talker model both immediately generates contextually grounded responses to hide the latency of an external \reasoner model and fluently integrates streamed \reasoner knowledge into its responses during inference. We curate a 290{,}571-example synthetic dataset spanning six domains and demonstrate that this task is learnable across seven widely used small language models ranging from 135M to 1.7B parameters\footnote{
Dataset:~\href{https://github.com/zenglhardt/convfill-dataset}{github.com/zenglhardt/convfill-dataset},\\Code and Models:~\href{https://github.com/vysri/conversational-infill}{github.com/vysri/conversational-infill}}. 
Our system implementation, \textit{ConvFill}, sustains millisecond-level time-to-first-response while closing the accuracy gap to within 6.3\% of the corresponding frontier \reasoner{} performance. In a live user study ($n=18$) with \talker deployments running on an Apple M2 SoC, participants rank ConvFill on par with frontier models overall, prefer it for retrieval-heavy tasks, and rate it significantly more responsive. These results show that conversational infill unlocks a new point on the latency--capability Pareto frontier, offering a practical path toward voice agents that are both responsive and highly capable.
\end{abstract}

\section{Introduction}
\begin{figure}[t!]
\centering
\includegraphics[width=\columnwidth]{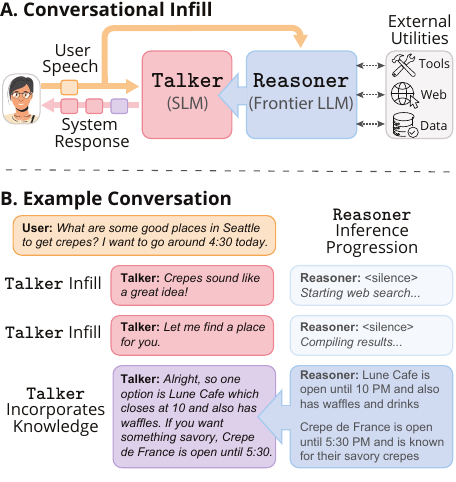} 
\caption{\textbf{Conversational infill.} \textbf{(A)} A lightweight \talker model (red) immediately starts responding while a \reasoner model (blue) performs higher-latency reasoning and tool calls. When the \reasoner responds, the \talker incorporates this information into subsequent responses (purple). \textbf{(B)} A conversation turn showing user speech, \talker response, and the background \reasoner inference progression. The third response phrase (purple) is conditioned on \reasoner context.}
\label{fig:fig1}
\vspace{-0.25in}
\end{figure}
Advances in large language models (LLMs) have enabled increasingly capable task-oriented systems for applications such as customer service, virtual assistants, tutoring, and interactive software agents. However, voice-based conversational models consistently lag behind their text-only counterparts~\cite{vera}. This gap stems from a fundamental tension: the techniques that enhance text-based model capabilities, such as multi-step reasoning, tool calls, and retrieval, are inherently iterative and time-consuming, while spoken interaction demands responses on a millisecond timescale to maintain a seamless user experience~\cite{humanlatency-avatarsystems}. These pressures pull in opposite directions: advanced capabilities improve answers but erode the responsiveness that conversational interaction depends on~\cite{tooluse-frontier-voice-models}.

To bridge this gap, we introduce \textbf{conversational infill}, a language model collaboration task in which a lightweight~\talker model interacts directly with the user, while a powerful~\reasoner model guides it with knowledge during inference. Our approach is inspired at a high level by the Talker-Reasoner concept presented in~\citet{talker-reasoner-christakopoulou}, but is distinct in formalization and implementation. The Talker-Reasoner division of labor allows the \talker to satisfy the millisecond-level responsiveness that spoken interaction requires while allowing the \reasoner to contribute its full capabilities without real-time constraints.

In our task, the \reasoner model is a frontier model able to handle multi-turn context and leverage reasoning and tool calls to generate knowledge chunks. These concise, non-conversational phrases prioritize the essential information needed to respond to a user utterance. 
The \talker model is a small (e.g., on-device) language model (SLM). The \talker model's primary role is to seamlessly transform knowledge chunks from the \reasoner into natural conversation. When no external knowledge is available, the \talker generates contextually grounded filler responses to hide \reasoner latency while maintaining conversational flow. For a task overview and representative interaction turn, see~\autoref{fig:fig1}.

Existing real-time voice systems fall into two categories. Cascaded systems combine separate automatic speech recognition (ASR), text-input language models, and text-to-speech (TTS), compounding latency and introducing gaps that break conversational flow. Full-duplex speech models~\cite{moshi, sesame_ai, lucy} generate audio directly, reducing latency by bypassing ASR and TTS steps. In comparison to cascaded approaches, full-duplex systems often use a single LLM backbone, sacrificing flexibility, while real-time constraints limit their size~\cite{audiobench-contrast}.

\textbf{ConvFill}, our end-to-end system implementing the conversational infill task, balances these tradeoffs. It keeps the flexibility of a cascaded design while avoiding the latency that usually comes with it. Because the user interacts only with the \talker, which receives \reasoner knowledge as it arrives, filler responses mask \reasoner inference time and the \talker never stalls waiting on the \reasoner. ConvFill's cascaded structure also keeps training tractable. The \talker takes transcriptions as input and therefore can be trained on text alone, without the resource-intensive audio modeling that full-duplex approaches require~\citep{kyutai-moshika-model-card, scaling_slms}. Lastly, ConvFill is not tied to a single choice of \talker or \reasoner model, meaning that any capable \talker or \reasoner can be slotted in independently.

We formalize the conversational infill task and demonstrate its feasibility and utility via an interactive end-to-end system. Concretely, we:

\begin{enumerate}[noitemsep, topsep=0pt, leftmargin=*, font=\bfseries]
\item Curate and release the ConvFill dataset, consisting of 290{,}571 rigorously validated training examples across six domains.
\item Show that the conversational infill task is learnable across seven SLMs spanning four model families (Qwen, Llama, Gemma, and SmolLM).
\item Demonstrate that \talker models retain millisecond-level time-to-first-response while closing the accuracy gap to within 6.3\% of the corresponding \reasoner performance. 
\item Build a voice-based, real-time ConvFill system deployed on an Apple M2 SoC and conduct a live evaluation $(n=18)$. Users rate ConvFill on par with frontier models for interaction quality and significantly better for perceived latency.
\end{enumerate}
\noindent To foster further work on model collaboration for responsive, capable voice systems, we release all of our artifacts: the ConvFill dataset, our generation and validation pipeline, the training and inference code for the end-to-end ConvFill system, and the fine-tuned weights for all seven \talker models.

\label{sec:prob-formulation}
\begin{figure*}[t]
    \centering
    \includegraphics[width=\textwidth]{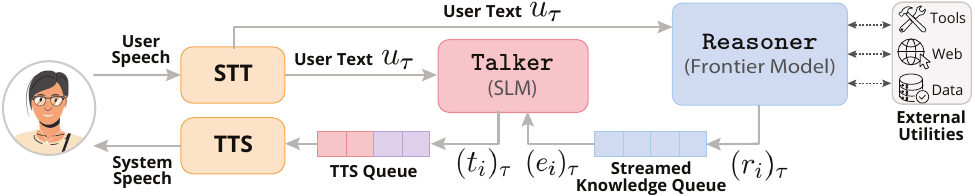} 
    \caption{\textbf{ConvFill System Architecture.} User speech is transcribed using STT and is sent concurrently to the \talker and \reasoner. The \talker generates filler responses (red) while waiting for the \reasoner (blue) to push knowledge chunks into the streamed knowledge queue. When there is \reasoner knowledge present in the streamed knowledge queue, the \talker ingests it and produces conditioned responses (purple). All \talker responses are passed to the TTS queue and processed to yield spoken output. } 
    \label{fig:inference-example}
    \vspace{-0.2in}
\end{figure*}

\section{Hiding Latency in Conversations}
Many techniques have been explored to compensate for latency in conversational systems. Prolonged silence, especially during real-time chat, results in interrupted conversational flow~\citep{flow-interruption-1, flow-interruption-2}. At a minimum, initial responses from conversational systems can acknowledge that the user has been heard, which is essential for responsive, natural-feeling conversation~\citep{conversational-turn-taking}. In the simplest case, a system responds with a fixed set of generic infill phrases, such as \textit{``I'm thinking.''} This is often extended by employing heuristics to select between different sets of infill phrases based on context~\citep{filler-2, lucy}. 

A more recent class of proposed approaches leverages dynamic infill methods that make use of multiple models at inference time. If the lower-latency model provides incorrect information, the higher-capability model corrects it with a more definitive answer. Such corrections, however, come at a cost: repeated revisions or inconsistencies---especially in spoken dialogue---have been shown to harm user trust, confidence, and engagement. Our architecture instead provides a consistent voice throughout full, multi-turn conversations by ensuring that the user only ever interacts with the \talker, which is trained to incorporate external knowledge rather than relaying it verbatim.

\section{Task Formulation} 
We propose a Talker-Reasoner architecture~\citep{talker-reasoner-christakopoulou} with a high-performance model as the \reasoner and a low-latency conversational model as the \talker. This results in a clear division of responsibilities between the two models. While the \reasoner uses tool calls, reasoning, and other advanced, yet high-latency, capabilities to address a user query, the \talker model must: 
\begin{enumerate}[noitemsep, topsep=0pt, leftmargin=*, font=\bfseries]
    \item produce conversationally contingent phrases to hide \reasoner latency \textit{when necessary}
    \item consume raw, inference-time information from the \reasoner \textit{when available} and transform it into fluent, contingent conversation
\end{enumerate}
In the sections that follow, we refer to our particular system-specific implementation of the conversational infill task as \textit{ConvFill}.

\subsection{Task Setup}
\label{subsec:task-setup}
In the following text, \talker refers to a lightweight small language model (SLM) capable of running on consumer-grade hardware. \reasoner refers to a high-performance model that can manage multi-turn context and engage with tools and external knowledge sources. We refer to a \textit{turn} as one exchange between a user and the \talker model. We use the notation $a \rightarrow b$ to mean that $a$ entails $b$~\cite{williams_broad-coverage_2018}. 

\subsubsection{\texorpdfstring{\talker}{Talker} Interface}
\label{subsec:talker-model}
Let $\tau$ denote the turn index in the conversation, where a conversation may consist of multiple turns. For turn $\tau$, $u_\tau$ is the user utterance at turn $\tau$. For this single turn, the \talker model responds in a sequence of phrases to form its full response, $T_\tau$, where $T_\tau = (t_1, t_2, \ldots, t_p)_\tau$.
$h_\tau$ denotes conversation history for turn $\tau$, with a lookback of $\ell$ turns:
    \[h_\tau(\ell) = (u_{\tau-\ell} \oplus T_{\tau-\ell}, \ldots, u_{\tau-1} \oplus T_{\tau-1})\]
    where $\oplus$ denotes concatenation.
Within a single turn $\tau$, let $i \in [1,\ldots,p]$ denote the phrase index.

The \talker{} model operates at the turn level, generating response $T_\tau$ phrase by phrase. At each index $i$ within the turn, the \talker{} conditions its generation on static and dynamic context. Static context includes conversation history $h_\tau(\ell)$, where $\ell$ is variable, and the current user utterance, $u_\tau$. Dynamic context is the knowledge stream $(e_1, \ldots, e_i)_\tau$ from the \reasoner up to position $i$, and phrases $(t_1, \ldots, t_{i-1})_\tau$ already generated in the current turn.

\subsubsection{\texorpdfstring{\reasoner}{Reasoner} Interface}
\label{subsec:reasoner-model-system}
The \reasoner{} interface asynchronously produces external knowledge for the \talker at inference time. It generates a stream of external knowledge chunks, $E_\tau = (e_1, e_2, \ldots, e_p)_\tau$. The $e_i$ may be either a knowledge chunk from the \reasoner, $r_i$, or a silence token, $s_i$.
The $r_i$ are semantically complete text phrases that convey the results of reasoning, retrieval, inference, tool calls, or other computation. Silence tokens $s_i$ are inserted dynamically if the \reasoner{} model has not sent a knowledge chunk after a predetermined time. During inference, the \reasoner has access to the full conversation history $h_\tau(\tau-1)$ and the current user utterance $u_\tau$ and uses these as context to generate each $r_i$. We assume that the \reasoner has a context window to store sufficient conversation history.

\subsubsection{Inference-Time Collaboration}
\label{subsec:inference-time-collaboration}
The \talker and \reasoner operate concurrently. The \reasoner streams knowledge chunks asynchronously, while the \talker continuously consumes from this stream. At each position $i$, the \talker observes the next element $e_i$. 

If $e_i = r_i$ (a knowledge chunk), the \talker should generate $t_i$ conditioned on $r_i$ in addition to the conversation history and its prior phrases. If $e_i = s_i$ (a silence token), which signals that the \reasoner has not produced output within a threshold duration, the \talker should generate $t_i$ conditioned only on the user utterance, conversation history, and prior phrases. Importantly $e_i = s_i$ and $e_i = r_i$ cannot occur at the same time. This is represented with notation in the equation below:
\begin{equation*}
\label{eq:inference-time}
\small 
\begin{cases}
    h_\tau(\ell) \oplus [u_\tau \oplus (t_1, \cdots, t_{i-1})_\tau] \oplus (r_i)_\tau \rightarrow t_i & \text{if } e_i = r_i \\
    h_\tau(\ell) \oplus [u_\tau \oplus (t_1, \cdots, t_{i-1})_\tau] \rightarrow t_i & \text{if } e_i = s_i
\end{cases}
\end{equation*}
This creates continuous output while respecting the \reasoner's variable latency and preventing the \talker from stalling while waiting for knowledge. We enforce the constraint that $|T_\tau| = |E_\tau|$.

\section{ConvFill Dataset} 
\label{sec:dataset}
To train the \talker models, we require a corpus of example conversations with user utterances paired with factual responses and corresponding conversational rephrasings. In addition, as specified in \autoref{sec:prob-formulation}, maintaining grounding solely in past and current inputs is crucial as no conversational response should reference information that would not be visible to the \talker at inference time. We create training examples that meet these requirements through a two-part process involving synthetic data generation paired with a series of phrase, turn, and conversation-level checks to detect structural and semantic violations. 

\noindent\textbf{Generation. }Conversations are structured as JSON objects containing a \texttt{conversation} array of turns. Turns carry a \texttt{user} utterance, a \texttt{thoughts} array beginning with 0--3~\standout{<sil>} silence placeholders followed by statements corresponding to \reasoner model phrases, and a \texttt{response} array with phrases corresponding to \talker responses. Entries in locations corresponding to~\standout{<sil>} placeholders are conversationally contingent filler phrases, while entries corresponding to \reasoner phrases incorporate the meaning of these phrases in a form and style consistent with the prior conversation. 

To ground conversations in realistic use cases, we first generate samples based on task-oriented dialogues from DSTC8~\cite{rastogi2020towards}, retaining the labeled user utterances and using the labeled system responses as the contents of the \reasoner \texttt{thoughts} array. This array is padded with~\standout{<sil>} placeholders at the beginning of each turn; \talker responses are generated conditioned on this array. To further expand the variety of conversational topics, we construct a second corpus consisting of randomly seeded topics in the domains of general advice, assistant queries, event planning, customer service, education, and medicine. Before writing conversations, we create a list of 1,000 unique conversation premise seeds per domain in order to increase conversation diversity. A conversation template consisting of 7--10 turns with 0--3~\standout{<sil>} placeholders per turn is created; all three streams are generated by an LLM according to this template, unlike in the DSTC8 case.

\noindent\textbf{Validation. }Despite significant effort tuning the prompts used to generate data, we found that current frontier models such as Claude Opus 4.6~\cite{claude_opus4.6} still frequently make errors such as subtly referencing future \texttt{thoughts} array contents in earlier \texttt{response} phrases. To prevent errors like this from contaminating the training dataset, we iteratively re-generate each conversation until it passes a strict four-stage validator. 

Candidate generations are first parsed to verify that they contain the appropriate number of turns and that matching \talker and \reasoner phrase pairs are present. Next, we detect ordering errors by computing pairwise BERTScore~\cite{zhang2020bertscore} between each entry in the \texttt{thoughts} array and each entry in the \texttt{response} array for each turn to ensure each entry relates most highly to the entry within its temporally aligned pair. Third, we utilize DeBERTaV3~\cite{he_debertav3} fine-tuned on MNLI~\cite{williams_broad-coverage_2018,noauthor_moritzlaurer/deberta-v3-base-mnli_nodate} to verify each grounded phrase pair exhibits entailment (see \autoref{appx:validation} for details). Finally, we verify that no proper nouns appear in any \texttt{response} field before first appearing in \texttt{user} or \texttt{thoughts} fields to prevent entity hallucinations.

\noindent\textbf{Composition. }The final ConvFill dataset consists of 8{,}443 conversations yielding 290{,}571 training examples: 111{,}552 silence-based filler examples and 179{,}019 knowledge-grounded examples in a JSONL format. The dataset is fully in English. The API cost of end-to-end dataset generation using Claude Opus 4.6 was approximately \$2{,}400, including retries due to validation failures. For prompt templates, validation details, and additional dataset statistics, see~\autoref{appx:dataset_generation}.

\section{Infill Model Training and Inference}

\subsection{Training Pipeline} 
We create a training pipeline that can be used to fine-tune models for the conversational infill \talker task. The pipeline is configured for each \talker model via control token sequences that delimit roles and turn boundaries. Most pretrained SLMs already define control sequences for the user role (\standout{[USER]}), assistant role (\standout{[ASST]}), and end-of-turn boundary (\standout{[END]}). For the infill task, we add a new control sequence, \standout{[KNOWLEDGE]}, and map it through each model's existing control-token template. Examples of expanded control sequences for each model and role can be found in \autoref{tab:boundary_tokens}. We add a new special token, \standout{<sil>}, to indicate cases in which \reasoner knowledge is not yet available.

\begin{figure}[ht]
\centering
\begin{prompt} 
\textcolor{blue}{
\standout{[USER]} (history) user utterance $\tau-1$  \standout{[END]}\\
\standout{[ASST]} (history) assistant utterance  $\tau-1$ \standout{[END]}}\\
\standout{[USER]} user utterance $\tau$  \standout{[END]}\\
\standout{[KNOWLEDGE]} external knowledge chunk ($e_i$) \standout{[END]} \\
\textbf{\standout{[ASST]} $t_1, \ldots, t_{i-1}$, \textcolor{red}{$t_i$ \standout{[END]}}}
\end{prompt}
\vspace{-5mm}
\caption{\textbf{Format for ConvFill training and inference.} During training, the model predicts the red text. The figure shows, in blue, a history lookback of $\ell=1$, as defined in \autoref{subsec:task-setup}.} 
\label{fig:convfill-prompt-format}
\vspace{-2mm}
\end{figure}

Training examples for \talker model fine-tuning are constructed as shown in \autoref{fig:convfill-prompt-format}, following the task description in \autoref{sec:prob-formulation}. Training examples contain a history lookback, shown in blue in \autoref{fig:convfill-prompt-format}, to ensure consistency across turns. Further history can be added in the same format if $\ell$ is increased. All training examples also contain the most recent user utterance, an external knowledge chunk, and, optionally, previous \talker phrases for the same turn, displayed in black in \autoref{fig:convfill-prompt-format}.

\subsection{Inference Pipeline}
\label{sec:inference}
We design ConvFill to implement the \talker-\reasoner architecture defined in \autoref{sec:prob-formulation}. We use a series of queues to implement ConvFill as shown in~\autoref{fig:inference-example}. The first queue buffers \reasoner knowledge chunks for ingestion by the \talker, and the second queue buffers \talker responses for ingestion by TTS. During ConvFill inference, a user utterance is transcribed and sent in parallel to both the \talker and \reasoner. The \talker performs greedy inference to process all available knowledge chunks and queue TTS responses. If the streamed knowledge queue is empty and the TTS queue is non-empty, the \talker waits until another knowledge chunk is available. However, if the TTS queue is empty, ConvFill dynamically produces a \standout{<sil>} token for the \talker to ingest, causing it to generate a conversationally contingent infill phrase. In practice, \standout{<sil>} tokens are inserted most frequently at the start of a conversation turn while waiting for the \reasoner to push a knowledge chunk into the streamed knowledge queue.

\subsection{\texorpdfstring{\talker}{Talker} and \texorpdfstring{\reasoner}{Reasoner} Models}
\label{subsec:eval-models} 
We fine-tune seven \talker models to perform the conversational infill task: Gemma 3 270M, Gemma 3 1B~\cite{gemma3}, Qwen3 0.6B~\cite{qwen3}, SmolLM2 135M, SmolLM2 360M, SmolLM2 1.7B~\cite{smollm2}, and Llama 3.2 1B~\cite{llama3}. This covers a range of parameter counts from 135~M to 1.7~B across 4 model families. Multiple sizes of SmolLM and Gemma are included to investigate scaling within model families. 

Per-model training configurations are available in \autoref{tab:talker-training-configs} and \autoref{appx:subsec:talker-base-models}. All \talker models are trained with a history lookback of size one ($\ell =1$; see \autoref{subsec:task-setup} for the definition). In our evaluations, we use three \reasoner models across different families: Claude Opus 4.7~\cite{claude_opus4.7}, GPT-5.5~\cite{gpt5.5-model_card}, and Gemini 3.1 Pro~\cite{gemini3.1_model_card}. The seven \talker and three \reasoner models yield 21 different ConvFill system configurations, which we evaluate across benchmarks and user studies. We convert all \talker models to the INT8 MLX format using MLX-LM~\cite{mlxlm2024} to enable benchmarking and evaluation directly on target hardware;~\talker model inference for all experiments in~\autoref{sec:benchmark-evaluation} and~\autoref{sec:user-evaluation} runs on an Apple MacBook Pro (M2 SoC, 16 GB of system memory).

\section{Benchmark Evaluation}
\label{sec:benchmark-evaluation}
\begin{table*}[t]
\centering
\scriptsize
\setlength{\tabcolsep}{1.5pt}
\renewcommand{\arraystretch}{1.25}
\begin{tabular*}{\textwidth}{@{\extracolsep{\fill}} >{\raggedright\arraybackslash}m{2.35cm} >{\raggedright\arraybackslash}m{5.55cm} >{\raggedright\arraybackslash}m{1.8cm} >{\raggedright\arraybackslash}m{1.85cm} >{\centering\arraybackslash}m{1.05cm} >{\centering\arraybackslash}m{1.05cm} >{\centering\arraybackslash}m{1.3cm} @{}}
\toprule
\textbf{Metric} & \textbf{Definition} & \textbf{Units} & \textbf{Method} & \shortstack[c]{\textbf{Single-}\\\textbf{turn}} & \shortstack[c]{\textbf{Multi-}\\\textbf{turn}} & \shortstack[c]{\textbf{Live}\\\textbf{Interaction}} \\
\midrule
Accuracy
  & Proportion of responses that correctly answer the question
  & \%
  & LLM Judge (QA)
  & \checkmark &  &  \\
\rowcolor{gray!8}
Conditional Accuracy
  & \talker{} accuracy excluding instances of incorrect \reasoner{} knowledge
  & \%
  & LLM Judge (QA)
  & \checkmark &  &  \\
Entailment
  & Ratio of \reasoner{} thoughts within each turn that entail the corresponding \textbf{non-filler} \talker{} phrase
  & \%
  & NLI Classifier
  & \checkmark & \checkmark &  \\
\rowcolor{gray!8}
Non-Contradiction
  & Ratio of \textbf{filler} \talker{} phrases within each turn that do not contradict \reasoner{} phrases
  & \%
  & NLI Classifier
  & \checkmark & \checkmark &  \\
Coverage
  & Degree to which \reasoner{} knowledge is reflected in the \talker{} response
  & Likert Scale (1--5)
  & LLM Judge
  & \checkmark & \checkmark &  \\
\rowcolor{gray!8}
Faithfulness
  & Degree of consistency between \reasoner{} and \talker{}
  & Likert Scale (1--5)
  & LLM Judge
  & \checkmark & \checkmark &  \\
Helpfulness
  & Binary indicator of whether \talker{} addresses the user's utterance in an appropriate manner
  & True/False (0--1)
  & LLM Judge
  & \checkmark & \checkmark &  \\
\rowcolor{gray!8}
Time to First Response (TTFR)
  & Time from transcription end to TTS queueing
  & ms
  & Live Logging
  &  &  & \checkmark \\
User Experience
  & Subjective ratings for Latency, Clarity, Fluency, Response Length, Coherence, Task Completion, Naturalness, Satisfaction
  & Likert Scale (1--5)
  & Survey (Likert)
  &  &  & \checkmark \\
\rowcolor{gray!8}
System Preference
  & Ordered ranking of systems according to preference
  & Rank (1--3)
  & Survey (Ranking)
  &  &  & \checkmark \\
\bottomrule
\end{tabular*}
\caption{\textbf{Evaluation metrics for ConvFill.} Check marks indicate which evaluation settings include each metric. Definitions are provided per-metric, describing what evaluation aspects each metric covers. Method indicates how the metric was measured. Metrics are measured across single-turn \& multi-turn benchmarks and live interaction.}
\label{tab:eval_axes}
\vspace{-0.2in}
\end{table*}

No single metric suffices to evaluate ConvFill: a system can answer accurately but express the answer unfaithfully, ground faithfully but omit critical information, or remain factually grounded while failing to address the user's intent. These failure modes are particularly difficult to disentangle in a dual-model system operating over an evolving dialogue, where errors can originate in the \reasoner, the \talker, or their interaction. We therefore construct a comprehensive evaluation spanning single-turn benchmarks, multi-turn benchmarks, and a live interaction study with human users, summarized in~\autoref{tab:eval_axes}. This section focuses on benchmark evaluations; live interaction results are presented in~\autoref{sec:user-evaluation}.

\subsection{Methods}
\label{subsec:benchmark-eval-methods}
For single-turn QA accuracy benchmarking, we compare four conditions: Base SLM (no conversational infill), ConvFill \talker, ConvFill \reasoner, and Frontier Model (no conversational infill). For NLI and LLM-as-a-Judge metrics in this section, we evaluate ConvFill \talker outputs only.
\subsubsection{Benchmark Metrics}
\label{subsubsec:benchmark-metric-defs}

\noindent\textbf{NLI Metrics. }Following terminology presented in~\citeauthor{williams_broad-coverage_2018}, we define turn-level notions of \textit{Entailment} and \textit{Non-Contradiction} for the conversational infill task. Since \standout{<sil>}-conditioned and \reasoner-knowledge-conditioned phrases serve distinct purposes, we evaluate them differently---\textit{Entailment} applies \textit{only} to \talker phrases in a turn that are generated from \reasoner knowledge. \textit{Non-Contradiction} applies \textit{only} to \talker phrases in a turn that are \standout{<sil>}-conditioned.

\textit{Entailment} measures the proportion of knowledge-conditioned \talker phrases that are entailed by their corresponding \reasoner phrase. This determines the \talker's ability to ground its responses in \reasoner-provided knowledge. \textit{Non-Contradiction} measures the proportion of \standout{<sil>}-conditioned \talker phrases that do not contradict \textit{any} of the \reasoner phrases in the turn. This measures the \talker's ability to generate non-disruptive filler phrases for a turn. For both metrics, we normalize by number of relevant phrases per turn to standardize across turns with varied phrase count.

\noindent\textbf{LLM-as-a-Judge Metrics. }Individual phrases generated by the \talker can score highly on NLI metrics, but when combined together still omit critical information, fail to address user intent, or remain incoherent at the turn level. We therefore complement phrase-level NLI with LLM-as-a-Judge metrics~\cite{zheng_judging_2023} that operate on the entire dialogue turn, motivated by prior work characterizing desirable model behavior in terms of helpfulness and faithfulness~\cite{ouyang_training_2022}. The judge metrics are derived from full conversational context (user utterance, \reasoner thoughts, \talker response, and optionally history lookback). Treating the \reasoner output as ground truth, these metrics distinguish between complementary failure modes: whether the \talker adequately covers the \reasoner's knowledge (Coverage, Faithfulness) and whether it appropriately addresses the user (Helpfulness).

\textit{Coverage} measures how much of the available \reasoner knowledge is reflected in the overall response, capturing omissions or under-specification, analogous to content-selection and recall-oriented evaluation in summarization~\cite{nenkova_evaluating_2004}. \textit{Faithfulness} measures whether the response introduces distortions, contradictions, or unsupported claims relative to \reasoner knowledge, capturing hallucination and misrepresentation errors~\cite{maynez_faithfulness_2020, precise_info_control}. Together, these metrics form a precision-recall view of grounding quality, separating omission failures from unsupported addition or distortion.

\textit{Helpfulness} captures the user-intent alignment component of instruction-following evaluation: whether the response meaningfully addresses the user's utterance, is coherent as a conversational turn, and is useful in context. Unlike broader notions of helpfulness that may implicitly include factual adequacy, our helpfulness metric isolates conversational usefulness from correctness with respect to the \reasoner knowledge, which is measured separately through Coverage and Faithfulness. Unlike Coverage and Faithfulness, which evaluate the \talker relative to the \reasoner, Helpfulness evaluates the \talker relative to the user. This isolates conversational capability from factual grounding. We express Helpfulness as a binary indicator, as human annotators in our pre-study pilot found a 1--5 scale difficult to apply consistently. Following prior work showing that thresholded Likert responses correlate more strongly with human judgments than direct binary questions~\cite{movva_annotation_2024, wang_improving_2025}, we elicit a Likert score from the LLM-as-a-Judge evaluator and binarize at a threshold of $\geq 3$.

LLM-as-a-Judge grading is performed by GPT-4o~\cite{gpt-4o} using the prompt in~\autoref{appx:subsec:llm-judge-prompt}. We observe sufficient inter-rater reliability between human graders, with a minimum Krippendorff's $\alpha$ of 0.67, and human--LLM agreement, with a minimum Kendall's $\tau_b$ of 0.55 and point-biserial $r$ of 0.69. Per-metric statistical analysis and the human grading rubric are provided in~\autoref{appx:subsec:infill-grading}. 

\subsubsection{Evaluation Conditions}
\label{subsubsec:benchmark-eval-conditions}
We run inference on text inputs only in this section. To simulate \reasoner latency, we prepend a random number of \standout{<sil>} tokens (0--3) before knowledge chunks during evaluation. We verify that this number of \standout{<sil>} tokens matches or exceeds the number generated in a real-world deployment of the ConvFill system (\autoref{tab:sil}). All SLM inference (Base SLM and \talker) for this section uses the setup specified in~\autoref{subsec:eval-models}. 

\begin{figure*}[t!]
\centering
  \vspace{0pt}
  \centering
  \includegraphics[width=0.95\linewidth]{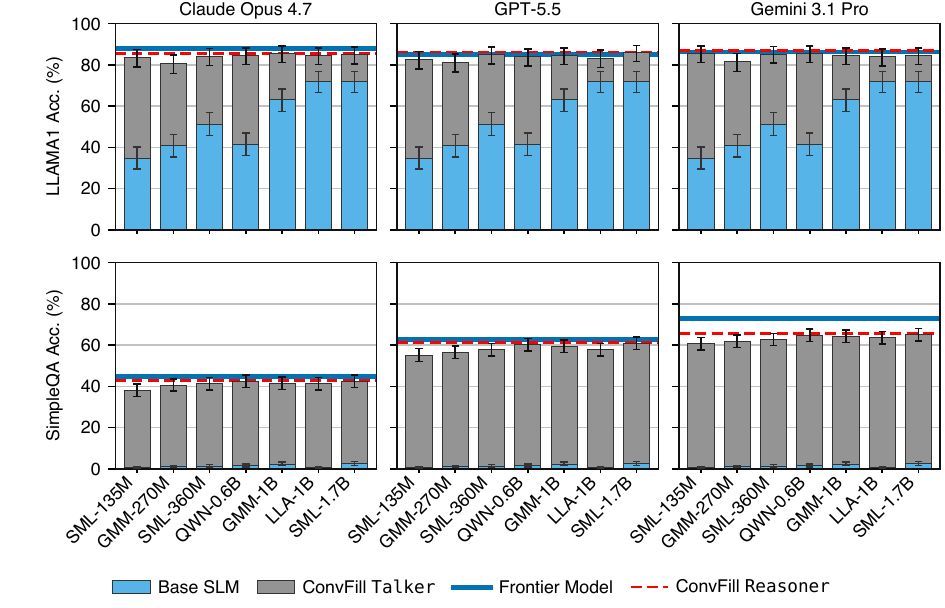}
\caption{\textbf{QA accuracy for model configurations across LLAMA1 and SimpleQA.} Each panel compares seven \talker models paired with one \reasoner. Bars show Base SLM and ConvFill \talker{} accuracies; horizontal lines mark Frontier model and ConvFill \reasoner{} accuracies. Error bars show 95\% confidence intervals. SML=SmolLM, GMM=Gemma 3, QWN = Qwen3, LLA=Llama. Full numerical results are in \autoref{appx:full-qa-results}.}
\label{fig:qa_results}
\vspace{-0.2in}
\end{figure*}

\subsubsection{Single-Turn Evaluations}
\label{subsec:single-turn-eval} 
We use two question-answering (QA) datasets for single-turn evaluation. SimpleQA~\cite{simpleqa-openai} consists of human-written, short-response questions spanning a broad range of factual domains; because examples are adversarially filtered during construction to retain only questions answered incorrectly by GPT-3.5~\cite{openai_introducing} or GPT-4~\cite{openai_gpt-4_2024}, it represents a set of intrinsically difficult factual QA examples. The LLAMA1 test set~\cite{google-llama1}, by contrast, consists of LLM-generated open-domain factual questions and is not filtered for failures of strong models, making it a comparatively easier factual QA benchmark.

For benchmarking with SimpleQA, we randomly sample 1,000 examples from the test split to form our evaluation set. Questions are prepared by following the SimpleQA dataset prompt and evaluation script~\cite{noauthor_openaisimple-evals_2026} and passed into the \talker and \reasoner models, simulating ConvFill. \reasoner responses are prepended with \standout{<sil>} tokens (see \autoref{subsubsec:benchmark-eval-conditions}) and are used as the basis for \talker generation. All \talker-generated phrases together make up the final response. 

Responses were graded for accuracy via the suggested GPT-4o judge model and evaluation prompt~\cite{noauthor_openaisimple-evals_2026}. Responses classified as~\textit{Not Attempted} and~\textit{Incorrect} are both considered to be incorrect for the purposes of tabulating accuracy metrics. Additional NLI and LLM-as-a-Judge metrics are calculated as indicated in~\autoref{tab:eval_axes}. LLAMA1~\cite{google-llama1} is a similarly formatted dataset that covers simpler general knowledge and elementary science; models were evaluated on the full 300-question test set with the same prompting and judge model as SimpleQA. 

\noindent\textbf{Conditions.}
We evaluate the following four conditions: Base SLM (the non-fine-tuned base SLM), ConvFill \talker (our fine-tuned model, fed by knowledge from the \reasoner at inference time), ConvFill \reasoner (the knowledge fed to the \talker during inference, graded independently), and, as a baseline, the Frontier model (the standalone frontier model corresponding to the ConvFill \reasoner, prompted directly for QA independent of ConvFill).

\begin{table*}[t]
\centering
\small
\setlength{\tabcolsep}{4pt}
\begin{tabular}{@{}l@{\hspace{10pt}}ccccc@{\hspace{16pt}}ccccc@{}}
\toprule
& \multicolumn{5}{c}{\textbf{LLAMA1}} & \multicolumn{5}{c}{\textbf{SimpleQA}} \\
\cmidrule(lr){2-6}\cmidrule(lr){7-11}
Model & Ent. & Non-C. & Cov. & Faith. & Help. & Ent. & Non-C. & Cov. & Faith. & Help. \\
\midrule
\rowcolor{gray!8}
SmolLM2 135M & 95.1 & 94.1 & 4.97 & 4.68 & 0.97 & 90.0 & 89.5 & 4.84 & 4.41 & 0.90 \\
Gemma 3 270M & 93.8 & 96.7 & 4.96 & 4.68 & 0.96 & 89.7 & 97.2 & 4.84 & 4.54 & 0.93 \\
\rowcolor{gray!8}
SmolLM2 360M & 95.7 & 94.8 & 5.00 & 4.84 & 0.99 & 95.4 & 94.4 & 4.91 & 4.68 & 0.94 \\
Qwen3 0.6B & 95.4 & 90.2 & 5.00 & 4.87 & 0.99 & 96.6 & 93.7 & 4.92 & 4.83 & 0.97 \\
\rowcolor{gray!8}
Gemma 3 1B & 96.3 & 86.8 & 5.00 & 4.86 & 0.99 & 95.6 & 92.2 & 4.91 & 4.72 & 0.96 \\
Llama 3.2 1B & 96.7 & 89.4 & 5.00 & 4.79 & 0.99 & 95.5 & 89.2 & 4.92 & 4.66 & 0.96 \\
\rowcolor{gray!8}
SmolLM2 1.7B & 95.8 & 98.1 & 5.00 & 4.88 & 0.99 & 97.0 & 95.8 & 4.92 & 4.88 & 0.98 \\
\specialrule{\heavyrulewidth}{0.8ex}{0.4ex}
& \multicolumn{5}{c}{\textbf{Everyday Conversations}} & \multicolumn{5}{c}{\textbf{MultiWOZ}} \\
\cmidrule(lr){2-6}\cmidrule(lr){7-11}
Model & Ent. & Non-C. & Cov. & Faith. & Help. & Ent. & Non-C. & Cov. & Faith. & Help. \\
\midrule
\rowcolor{gray!8}
SmolLM2 135M & 92.3 & 95.6 & 4.99 & 4.87 & 1.00 & 83.9 & 85.9 & 4.83 & 4.57 & 0.91 \\
Gemma 3 270M & 85.1 & 96.5 & 4.99 & 4.77 & 0.99 & 82.9 & 81.0 & 4.85 & 4.55 & 0.92 \\
\rowcolor{gray!8}
SmolLM2 360M & 91.4 & 97.9 & 5.00 & 4.91 & 1.00 & 82.8 & 84.5 & 4.86 & 4.68 & 0.94 \\
Qwen3 0.6B & 92.0 & 96.7 & 5.00 & 4.88 & 1.00 & 83.0 & 81.8 & 4.90 & 4.74 & 0.94 \\
\rowcolor{gray!8}
Gemma 3 1B & 84.8 & 98.2 & 5.00 & 4.91 & 1.00 & 82.3 & 81.2 & 4.85 & 4.52 & 0.92 \\
Llama 3.2 1B & 87.8 & 97.7 & 5.00 & 4.86 & 1.00 & 84.5 & 81.3 & 4.87 & 4.60 & 0.93 \\
\rowcolor{gray!8}
SmolLM2 1.7B & 85.1 & 99.1 & 5.00 & 4.97 & 1.00 & 83.6 & 82.8 & 4.89 & 4.74 & 0.95 \\
\bottomrule
\end{tabular}
\caption{\textbf{Compact NLI and GPT-4o judge metrics for QA datasets (top) and sampled multi-turn datasets (bottom).} QA values are averaged across the three hosted \reasoner{} models. Entailment (Ent.) and Non-Contradiction (Non-C.) are percentages; Coverage (Cov.), Faithfulness (Faith.), and Helpfulness (Help.) use the scales described in \autoref{tab:eval_axes}. Full results by \reasoner model and confidence intervals are in \autoref{appx:full-qa-results}.}
\label{tab:metrics_stacked}
\vspace{-0.2in}
\end{table*}


\subsubsection{Multi-Turn Evaluations}
\label{subsec:multi-turn-eval}
Multi-turn conversations pose an additional challenge because each response depends on the dialogue state established by previous turns; single-turn QA evaluations alone cannot capture failures in context tracking, referent resolution, consistency, or error accumulation across a conversation. For multi-turn evaluation, we use two datasets that differ substantially in conversational complexity. MultiWOZ~\cite{multiwoz} contains human--assistant task-oriented dialogues that average 13.5 turns and require tracking evolving user goals, constraints, and decisions across often multi-domain interactions. In contrast, Everyday Conversations~\cite{everydayconversations2024} consists of synthetic 3--4-turn conversations with simpler structure, making it a comparatively easier conversational continuity test.

Unlike in \autoref{subsec:single-turn-eval}, where we simulate the entire ConvFill system during evaluation, here, we treat the original dataset \textit{assistant} text at each turn as the \reasoner knowledge by splitting on sentence boundaries and prepending~\standout{<sil>} tokens. In order to account for history lookback, outputs are generated sequentially starting from the first turn of dialogue with each subsequent \talker response depending on the user field, simulated \reasoner knowledge, and the previous turn \talker response. We repeat this for all turns in the conversation. 

For each ConvFill-simulated conversation constructed from the 1,000-conversation MultiWOZ test set, we select a random turn for grading, excluding the first and last turns (which are usually greetings or closings). The corresponding \talker-generated part of the turn is then graded on the metrics indicated in~\autoref{tab:eval_axes}. This creates 991 graded turns, as conversations of two turns or less are implicitly excluded. We repeat this same process for the test set of Everyday Conversations, yielding 119 graded turns. 

\noindent\textbf{Conditions.}
We only evaluate the ConvFill \talker model output for this scenario, as all other multi-turn inputs directly use dataset contents or are generated by previous \talker inference turns.

\subsection{Results}
\label{subsec:benchmark-results}
\autoref{fig:qa_results} displays the accuracy of the ConvFill \talker model on the single-turn QA task relative to the corresponding Base SLM, \reasoner, and Frontier model. Since \reasoner knowledge chunks enable \talker performance gains, the accuracy of each \reasoner represents a ceiling for the corresponding \talker model performance. We see consistently across both SimpleQA and LLAMA1 benchmarks that ConvFill \talker models perform within 0\%--6.3\% of their corresponding \reasoner models. ConvFill \talker models, across \reasoner models, demonstrate gains as high as 51\% on LLAMA1 and 63.4\% on SimpleQA, relative to the corresponding base SLMs.

For the LLAMA1 dataset, the standalone \reasoner accuracy relative to Frontier-model-only accuracy is 2.3\% less for Claude, 0.7\% greater for Gemini, and 1.3\% greater for GPT-5.5. For the SimpleQA dataset, the standalone \reasoner accuracy relative to Frontier-model-only accuracy is 2.0\% lower for Claude, 7.4\% lower for Gemini, and 1.4\% lower for GPT-5.5. Across these, we observe that the \reasoner-Frontier gap is roughly 2\%, with the notable exception of Gemini 3.1 Pro, for which the gap is almost triple that of the other configurations on the SimpleQA task.

We display calculated NLI and LLM-as-a-Judge metrics in~\autoref{tab:metrics_stacked} for all four datasets. To determine the significance of trends with respect to \talker parameter size, we utilize the Benjamini–Yekutieli False Discovery Rate correction~\cite{benjamini_control_2001} to control for false positives while accounting for potential dependencies across multiple hypothesis tests. Full significance test results are shown in~\autoref{tab:family_scaling_comparison}.

On the two QA datasets, we observe a significant positive correlation between Base SLM parameter count and Accuracy, confirming that larger models have stronger underlying question-answering capabilities prior to any ConvFill system integration.
This baseline result aligns with established scaling intuitions and provides a foundation for interpreting downstream ConvFill system trends.
On the more challenging SimpleQA dataset, significant positive scaling trends emerge across nearly all metrics (Accuracy, Entailment, Coverage, Faithfulness, and Helpfulness), indicating that model size meaningfully improves response quality when the task places real demands on model capability.
On the easier LLAMA1 dataset, Accuracy, Entailment, and Coverage are uniformly high across all model sizes, leaving little room for scaling trends to emerge.
Positive trends are nonetheless still observed in Faithfulness and Helpfulness, suggesting these metrics remain sensitive to scale even when general correctness is largely saturated.

A similar pattern appears in the multi-turn evaluations. In MultiWOZ, we observe significant positive scaling trends for Coverage, Faithfulness, and Helpfulness, indicating that larger \talker models are better able to use \reasoner knowledge in longer, task-oriented dialogues. In contrast, no significant scaling trends emerge on Everyday Conversations. This divergence likely reflects the differing complexity of the two datasets and mirrors the single-turn contrast between SimpleQA and LLAMA1: scaling trends are most visible when the benchmark is challenging enough that model capability, rather than task saturation, limits performance.

Non-Contradiction is the main exception to the otherwise positive performance scaling with respect to \talker model size. It shows no clear relationship with \talker model size on the two single-turn QA datasets or on Everyday Conversations; on MultiWOZ it shows a negative correlation with model size. We turn to within-family comparisons to examine this unexpected trend with additional nuance. Across the three SmolLM2-derived \talker models, scaling trends are broadly consistent with the overall parameter count results, with the notable exception of Non-Contradiction: a significant positive correlation between size and Non-Contradiction is observed on both QA datasets, a reversal of the cross-family trend. The Gemma 3 family shows the opposite pattern, with larger models exhibiting worse Non-Contradiction on QA datasets. We discuss these observations further in \autoref{sec:discussion}.

\section{User Evaluation}
\label{sec:user-evaluation}
Rather than evaluating perceived metrics from static transcripts or recordings, we build a full end-to-end, interactive ConvFill system prototype to enable users to interact with it directly. We conduct a user study (18 participants: 10 male, 8 female, ages 20--63). The study included a live interaction portion in which participants conversed with voice agents both with and without the \talker-\reasoner architecture, and a grading portion in which participants rated Coverage, Faithfulness, and Helpfulness of samples to generate a validation corpus for LLM-as-a-Judge evaluation.
\subsection{Methods}
\label{subsec:methods-user-evaluation}
We evaluate three conversational system configurations: Base SLM, Frontier, and ConvFill. In the Base SLM configuration, participants conversed directly with a small language model prompted for conversational interaction. In the Frontier configuration, participants conversed directly with a frontier model prompted for conversational interaction. In the ConvFill configuration, participants conversed with the full ConvFill system, in which a fine-tuned \talker model generated the user-facing response while receiving streamed knowledge from a \reasoner model. Llama 3.2 1B, Qwen3 0.6B, and Gemma 3 270M were used as Base SLMs, while Claude Opus 4.7 and GPT-5.5 were used as Frontier models. The ConvFill counterparts for these models were used as \talker and \reasoner models in this evaluation, as described in~\autoref{subsec:eval-models}.

\noindent\textbf{Study Tasks. }In this evaluation, we augment standard chat to include two \textit{out-of-distribution tasks} that do not appear in our fine-tuning data: retrieval-augmented generation (RAG)~\cite{lewis_retrieval-augmented_2021}, and tool use through Model Context Protocol (MCP)~\cite{model-context-protocol}. We refer to the three tasks as Direct, RAG, and MCP. In the Direct task (standard chat), participants completed one of three task-oriented dialogue scenarios without external retrieval or tool access. In the RAG task, a database was connected to each Base SLM, \reasoner, or frontier model; participants completed scenarios requiring database information. In the MCP task, participants queried an email inbox via MCP.\footnote{For MCP, the \textit{Base SLM} configuration was omitted because the model could not make structured MCP calls.}

\noindent\textbf{Interaction Conditions.}
Participants each completed all three task types---Direct, RAG, and MCP---following a Latin Square order~\cite{mackenzie2024human}. Within each task, the order of system configurations (Base SLM, Frontier, or ConvFill) was randomized; the MCP configuration only featured Frontier and ConvFill configurations due to the Base SLM's inability to make MCP calls.

Participants were instructed to converse with each system for sufficient time to form an opinion, with a suggested minimum of 4 turns. After each interaction, participants rated each system on a 5-point Likert scale using the metrics in~\autoref{tab:user-study-ratings}. After completing all interactions for a task, participants also ranked the systems in order of preference. Full details are provided in~\autoref{appx:subsec:condition-assignment}.

\noindent\textbf{Real-Time Interactive System. }Each ConvFill live interaction is conducted with an end-to-end interactive prototype fully implementing the ConvFill system. We enable voice-based interaction using faster-whisper~\cite{fasterwhisper} for automatic transcription and the macOS Speech Synthesis Manager~\cite{noauthor_speech_nodate} for text-to-speech. The six Base SLM--Frontier LLM pairs were uniformly assigned across the 18 participants, ensuring each pair appeared three times. Pairs were held constant within each user session. We ran Base SLM and ConvFill~\talker models as specified in~\autoref{subsec:eval-models}; additional implementation details are provided in~\autoref{appx:subsec:live-interaction-setup} and the demo software in the project repository.

\noindent\textbf{Metrics. }We adopt eight metrics for evaluation of user perception. \textit{Latency} refers to the user perception of whether the system responded fast enough to maintain natural conversation. Following~\citeauthor{munro1995_clarity_fluency}'s intelligibility and comprehensibility metrics, we define \textit{Clarity} as the degree to which the user could understand the words the system said and \textit{Fluency} as how well the users could understand ideas expressed by the system. 
\textit{Response Length} refers to how appropriate the user felt the length of the response was. \textit{Coherence} measures whether sentences relate to each other and progress logically~\citep{barzilay2005_coherence}. \textit{Task Completion} refers to whether system responses provided information needed to complete the given task~\citep{mohammadi2025_task_completion}. \textit{Naturalness} refers to whether the language used felt appropriate for the dialogue context~\citep{danieli1996_naturalness}. \textit{Satisfaction} refers to how satisfied users were with their conversation with the system. We give each of the aforementioned metrics names so that we can refer to them; during the study, participants were only given \textit{descriptions} of the metrics to refer to (see \autoref{appx:user-study-instructions}).

\noindent\textbf{Infill Grading. }Participants completed an infill grading task in which they were given a user utterance, a set of external knowledge chunks, and a set of conversational responses. They were asked to grade Coverage and Faithfulness on 1--5 scales and Helpfulness as a binary judgment. Grading results were collected and used for correlation with LLM-as-a-Judge ratings mentioned in \autoref{sec:benchmark-evaluation}. For details, see \autoref{appx:subsec:infill-grading}.

\subsection{Results} 
\label{subsec:user-study-results}
Across all task and system conditions, the study collected 144 interactions comprising 635 total turns.

\begin{table}[t]
\centering
\small
\begin{tabular}{@{}lccc@{}}
\toprule
\textbf{Metric} & \textbf{Base SLM} & \textbf{Frontier} & \textbf{ConvFill} \\
\midrule
Latency           & $3.44 \pm 1.25$ & $3.46 \pm 1.28$ & $4.24 \pm 0.85$ \\
Clarity           & $4.53 \pm 0.84$ & $4.81 \pm 0.48$ & $4.69 \pm 0.64$ \\
Fluency           & $3.69 \pm 1.45$ & $4.83 \pm 0.42$ & $4.56 \pm 0.72$ \\
Resp. Length   & $3.00 \pm 1.43$ & $4.07 \pm 0.91$ & $3.94 \pm 1.05$ \\
Coherence   & $2.94 \pm 1.16$ & $4.72 \pm 0.46$ & $4.28 \pm 1.32$ \\
Task Comp.   & $2.33 \pm 1.49$ & $4.35 \pm 0.89$ & $4.31 \pm 0.84$ \\
Naturalness  & $3.61 \pm 1.05$ & $4.41 \pm 0.74$ & $3.83 \pm 1.08$ \\
Satisfaction      & $2.03 \pm 1.03$ & $3.80 \pm 1.05$ & $3.81 \pm 1.03$ \\
\bottomrule
\end{tabular}
\caption{\textbf{Likert ratings per system aggregated across tasks.} Mean $\pm$ std. dev. are reported. Base SLM has a sample size of $N=36$, and Frontier and ConvFill have a sample size of $N=54$. Metrics are described in \autoref{subsec:methods-user-evaluation}.}
\label{tab:user-study-ratings}
\vspace{-0.2in}
\end{table}

\noindent\textbf{User Ratings. }Summarized Likert rating results are in \autoref{tab:user-study-ratings}. We compare Frontier and ConvFill ratings with two one-sided tests (TOST) and a non-parametric Wilcoxon rank-sum test (see \autoref{appx:subsec:tost-details}). ConvFill was rated equivalently to the Frontier system on Clarity, Fluency, Response Length, Coherence, Task Completion, and Satisfaction ($p < 0.01$ for all, see \autoref{appx:subsec:tost-details}). ConvFill exceeded the Frontier configuration on Latency, but fell below the Frontier configuration on Naturalness: some users who were not used to models using filler phrases scored this metric lower. The Base SLM condition was rated significantly below ConvFill and Frontier across all metrics.

\noindent\textbf{System Rankings. }Rankings were pooled across all tasks, yielding $N = 54$ observations per system, except for the Base SLM, which only has $N = 36$ observations due to inability to make MCP calls. Users ranked the models for each task (1: best, 3: worst for Direct and RAG; 1: best, 2: worst for MCP) as shown in~\autoref{tab:user-study-ranking}. Rank-1 tests show no preference for ConvFill vs. Frontier on Direct and MCP tasks, but ConvFill was strongly preferred for RAG tasks (\autoref{appx:subsec:test-rankings}). Base SLM was ranked last by 15/18 users on Direct and RAG tasks and was excluded in the MCP task.

\begin{figure}[t]
\centering
\includegraphics[width=\columnwidth]{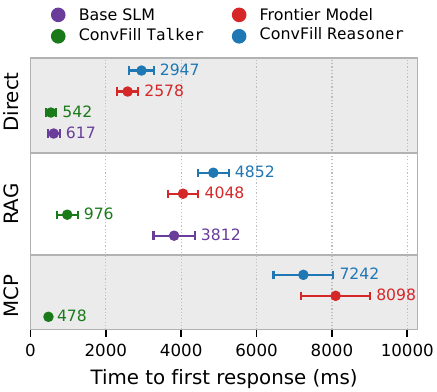}
\caption{\textbf{TTFR Analysis per Task.} Horizontal shaded regions indicate each task, while labeled points with 95\% CI error bars indicate latency. ConvFill \talker TTFR matches or improves on Base SLM TTFR despite ConvFill \reasoner and Frontier models demonstrating similar TTFR across tasks. Full statistics are in \autoref{tab:latency}.}
\label{fig:model-latency-plot}
\vspace{-0.2in}
\end{figure}

\begin{table}[t]
\centering
\small
\begin{tabular}{llccc}
\toprule
\textbf{Task} & \textbf{System} & \textbf{Rank 1} & \textbf{Rank 2} & \textbf{Rank 3} \\
\midrule
\multirow{3}{*}{Direct}
  & Base SLM   & 0  & 3  & 15 \\
  & Frontier    & 10 & 7  & 1  \\
  & ConvFill   & 8  & 8  & 2  \\
\midrule
\multirow{3}{*}{RAG}
  & Base SLM   & 1  & 2  & 15 \\
  & Frontier    & 5  & 11 & 2  \\
  & ConvFill   & 12 & 5  & 1  \\
\midrule
\multirow{2}{*}{MCP}
  & Frontier    & 7  & 11 & --  \\
  & ConvFill   & 11 & 7  & --  \\
\bottomrule
\end{tabular}
\caption{\textbf{Rank distribution by system and task.} Participant counts; $N=18$ per task. Rank 1 = best. Overall, participants rate ConvFill better than or generally on par with the Frontier model configuration.}
\label{tab:user-study-ranking}
\vspace{-0.3in}
\end{table}
\noindent\textbf{Latency \& Filler Analysis. }We analyze the time-to-first-response (TTFR) of all four configurations across all tasks in \autoref{fig:model-latency-plot}. We define \textit{TTFR} as the time from receipt of the user's transcribed message to the first generated full sentence, rather than the first token. We choose this granularity because it reflects when a complete sentence is ready to be voiced by the TTS system. We observe that all three tasks (Direct, RAG, and MCP) exhibit distinct, non-overlapping TTFR distributions, with ConvFill \reasoner and Frontier configurations exhibiting similar TTFRs across tasks. 
ConvFill \reasoner TTFR varied substantially, from a mean of 2{,}947\,ms on Direct tasks to 7{,}242\,ms on MCP tasks, due to latency of context retrieval and tool calls. Despite this, the ConvFill \talker filler responses remained consistently fast across tasks (Direct: 542\,ms, RAG: 976\,ms, and MCP: 478\,ms), with per-sample speedups averaging 7.4$\times$, 9.2$\times$, and 19.1$\times$ relative to the \reasoner{} response as shown in \autoref{appx:tab:speedup}. ConvFill emitted more filler tokens as \reasoner{} TTFR increased, compensating for varied \reasoner{} latencies across tasks (Direct: $\mu=2{,}947$\,ms, RAG: $\mu=4{,}852$\,ms, MCP: $\mu=7{,}242$\,ms). ConvFill \talker TTFR was similar to the Base SLM for Direct tasks (542\,ms vs.\ 617\,ms), but much lower for RAG tasks (976\,ms vs.\ 3{,}812\,ms), as the Base SLM incurs the burden of processing retrieved context while ConvFill offloads this to the \reasoner.

\section{Discussion}
\label{sec:discussion}
\noindent\textbf{Dynamic Fillers for Perceived Responsiveness.}
ConvFill substantially reduces both actual and perceived latency relative to frontier models, while retaining their accuracy on QA tasks. ConvFill consistently delivers millisecond-level-TTFR responses across three different task types with \reasoner latencies. This benefit is especially pronounced for tasks involving external resources, due to the inevitable access time to query external databases or email servers via third-party providers; particularly, these latencies are not reduced via model-level optimization. ConvFill dynamically adapts to variable \reasoner{} latency by inserting additional \standout{<sil>} tokens, emitting more \talker filler sequences as task complexity grows. This adaptive filling allows a single \talker{} to handle a range of latency conditions and variable infill lengths without task-specific TTFR tuning. User study participants' perceived latency ratings reflect these measurements: ConvFill received the strongest latency scores of any configuration, demonstrating alignment between TTFR measurements and subjective experience.

\noindent\textbf{Reducing Local Inference-Time Burden.}
ConvFill's measured \talker TTFR is equal to or lower than that of the corresponding base SLMs across all tasks. Additionally, assisted by the \reasoner model, the \talker recovers almost all of the frontier-model-only performance while maintaining this TTFR.
Simply using a small on-device model does not solve the TTFR problem---particularly in the RAG task, the base SLM infers almost as slowly as the frontier model despite being identical in size and architecture to the \talker. This is because the unassisted base SLM must independently process the full retrieved context, while the \talker gets phrase-based, condensed knowledge from the \reasoner model. Due to this modularity, the ConvFill architecture also reduces the inference-time token-processing burden on the \talker model, enabling it to remain responsive in conversation.

\noindent\textbf{Accuracy and Frontier Model Parity.} Accuracy results demonstrate that a \talker can faithfully incorporate \reasoner knowledge: the end-to-end ConvFill system approaches frontier model-only accuracy on QA tasks. There are two points where ConvFill performance could suffer relative to the unmodified Frontier Model baseline: prompting changes that induce the phrase-based \reasoner output format could hurt task performance; and \reasoner knowledge could be lost or corrupted as the \talker integrates it into conversation.

As illustrated in~\autoref{sec:benchmark-evaluation}, \talker accuracy on QA tasks generally approaches the performance ceiling set by the corresponding Frontier Model baseline. The one exception to this occurs with the Gemini 3.1 Pro~\reasoner on SimpleQA, where \talker performance trails baseline performance by an average of 9.8\%. Examining the ConvFill \reasoner results more closely indicates that this anomalous behavior results from degradation in frontier model performance when switched to the \reasoner output format. Thus, the discrepancy is better characterized as model-specific sensitivity of Gemini 3.1 Pro to the ConvFill \reasoner prompting context, rather than as a failure mode introduced by the \talker component itself.

This observation should not be discarded as an outlier; it is consistent with prior work showing that seemingly minor prompt choices such as formatting or example ordering can produce substantial, model-dependent performance differences~\citep{zhuo_prosa_2024, lu_fantastically_2022, sclar_quantifying_2024}. Our evaluation uses a single generic prompt structure across \reasoner models rather than tuning prompts separately for each model, providing a conservative estimate under a uniform deployment setting. In application-specific deployments, further task- and model-specific prompt adaptation could verify \reasoner robustness and provide an additional path to improve ConvFill performance.

\noindent\textbf{Task Difficulty and Metric Saturation.}
Benchmark results reveal a more nuanced picture of what the conversational infill task necessitates beyond QA accuracy alone. Grounding metrics such as Entailment and Coverage are consistently strong across model sizes and datasets, suggesting that \talker models reliably learn to incorporate and reflect \reasoner knowledge in their responses. Faithfulness and Helpfulness, which capture subtle aspects of response quality, namely whether the \talker distorts meaning or fails to address the user's intent, show greater variance across models and are more sensitive to task difficulty, with the MultiWOZ dataset yielding notably lower scores than the QA and Everyday Conversations datasets. This suggests that while the mechanical task of knowledge incorporation is learnable across all of the model sizes we evaluate, producing responses that are accurate in representation and appropriate in conversational context remains a more difficult and discriminating challenge. On sufficiently easy tasks, metric saturation obscures scaling behavior entirely, suggesting that benchmarks for ConvFill-type systems should be calibrated to a difficulty level where model capability is the limiting factor rather than the task ceiling.

\noindent\textbf{\talker Model Scaling.}
Accuracy, Coverage, Faithfulness, and Helpfulness all scale positively with model size, particularly on more challenging samples where performance is not saturated. Entailment scales on single-turn QA benchmarks but not on multi-turn benchmarks. We attribute this to the NLI checker being unable to robustly handle multi-turn nuances, further motivating our choice to evaluate with additional LLM-as-a-Judge metrics. The Non-Contradiction metric demonstrates opposing trends between SmolLM and Gemma model families~\autoref{tab:family_scaling_comparison}. This suggests that other factors, such as model architecture, training methods, and dataset composition, play a larger part in performance on this metric. 

\noindent\textbf{Live Interaction Takeaways.}
Conducting a live end-to-end evaluation enables users to assess the full ConvFill system directly under scenarios representative of real-world use.
The live setting also adds two forms of realism beyond the benchmark conditions.
First, it introduces two task types absent from training: retrieval-augmented generation and tool use through MCP, leaving the Direct task as the only condition close to the training distribution.
Second, it runs the full system on real users' unscripted interactions, with participants given only high-level guidance on potential topics and queries (\autoref{sec:user-evaluation}).
Under live conditions, results follow the general trend of the earlier benchmarks, with ConvFill and Frontier configurations significantly exceeding the Base SLM across metrics. ConvFill is statistically equivalent to the Frontier configuration on all metrics except for Latency, where it exceeds the Frontier configuration and Naturalness, where it trails (\autoref{appx:subsec:tost-details}).

In aggregate, users rank ConvFill on par with the Frontier configuration for the Direct and MCP tasks and prefer it outright for RAG as shown in~\autoref{tab:user-study-ranking}, suggesting that better responsiveness is preferable, even at the cost of some naturalness. The RAG and MCP tasks are also the two cases where the \reasoner must issue a query and wait to receive and encode an external result before responding, incurring the highest latency in our system. The clearest user preference appears on RAG rather than on MCP, even though MCP carries the longest \reasoner delays.
These longer delays produce more infill phrases (2.35 vs 1.34 per turn on average) before the \reasoner response arrives (\autoref{tab:sil}).
A possible explanation is that ConvFill introduces the most benefit during medium-length inference delays, and that at higher infill counts it may begin to detract from the experience relative to silence.

\noindent\textbf{Latency-Adaptive Behavior.}
The naturalness-responsiveness tradeoff suggests that simply filling all moments of silence with infill may not be optimal. Instead, a future ConvFill system could anticipate how long the \reasoner is likely to take and adapt accordingly. When a fast response is expected, ConvFill could hold off on non-grounded infill and wait briefly to answer from \reasoner knowledge. For longer delays, ConvFill could dynamically generate pauses when indicated by \talker filler content (e.g. \textit{``Let me take a moment to check your inbox\ldots{}''}), rather than sustaining constant filler until the response arrives. Tuning infill for context and expected latency in this way, rather than applying it uniformly, offers a path to preserving the responsiveness users value while
recovering the naturalness that filler can cost when in excess.

\noindent\textbf{Development Feasibility.} We demonstrate that the conversational infill task is learnable across a diverse set of small model architectures and parameter scales, from 135M to 1.7B parameters, spanning widely used Gemma, Qwen, SmolLM, and Llama families. Resource requirements for fine-tuning range from 2.8--48.9 GPU hours on an RTX 6000, with a total estimated cost of \$133.70 to train all 7 \talker models. Full, per-model breakdowns of cost and GPU hours are found in \autoref{tab:talker-training-configs}. Overall, adopting ConvFill for new application scenarios or \talker-\reasoner pairings does not require large-scale training infrastructure and can be done cheaply on previous-generation hardware.

We generated and verified our synthetic dataset at a one-time cost of \$2{,}400. This cost is due to both the size of the dataset and the sample-generation retries needed to meet strict validation requirements. 

\noindent\textbf{Target Devices.}
Our evaluation is representative of real deployment in both target hardware and model precision. We benchmark and run our live interaction study on a laptop with an Apple M2 SoC and 16 GB of memory, using the same INT8-quantized \talker models suitable for deployment, with graph sizes ranging from 136 MB to 1.69 GB (\autoref{tab:convfill-talkers-finetuned}). Measured latency, reported accuracy, and grounding metrics therefore reflect realistic deployment conditions~\cite{laskaridis_melting_2024} and illustrate the feasibility of deployment across a wide range of existing similarly equipped consumer and edge devices. Smartphones are an especially natural target: the \talker weights occupy only a fraction of the RAM in recent phones, whose SoCs include dedicated neural hardware with native support for INT8 inference~\cite{tan_mobilequant_2024}. This makes conversational infill a practical approach to bridge the latency-capability tradeoff between frontier-level intelligence and real-time responsiveness for applications like conversational digital assistants. 

\noindent\textbf{Decoupling Model Behavior and Capability.} The broader implication of the ConvFill architecture is a clean separation between the model responsible for conversational behavior (\talker) and the model responsible for advanced capabilities and reasoning (\reasoner). Decoupling the tasks in this way presents several advantages relative to prior conversational AI approaches.

First, it makes the \talker independently trainable. This enables optimizing for natural, task-appropriate spoken interaction without incurring the high cost of directly fine-tuning frontier text or audio models~\citep{kyutai-moshika-model-card, finetuning-costs-llms, finetuning-lr}. This also enables task-specific adaptation while avoiding the possibility of fine-tuning-induced performance degradation~\citep{finetuning-cat-forgetting}. Beyond this, we envision use cases in which the \talker can be adapted to enable stylistic variations or configurable conversational behaviors.

Second, this makes the ConvFill system modular---we show that \reasoner models can be freely swapped without retraining or altering the prompting of \talker models. This facilitates leveraging the variety of pre-existing tools and data sources already integrated with frontier-class models as drop-in \reasoner replacements without re-engineering or tuning them for real-time latency and conversational behavior. 

\noindent\textbf{Interaction Consistency Across Upgrades.} A longstanding challenge in deploying conversational systems or chatbots is keeping conversational behavior consistent across model iterations and upgrades, as users recognize these changes and often respond negatively~\cite{freitas_lessons_2025}. Maintaining a consistent customized~\talker model across \reasoner updates could enable model service providers to more easily maintain a consistent user experience. This possibility is especially attractive given the rapidly changing landscape of frontier model performance and token costs~\cite{chen_frugalgpt_2023}.

\section{Related Work}
Recent work has divided speech and reasoning across models to balance responsiveness and intelligence. Mini-Omni-Reasoner~\cite{miniomnireasoner-xie} applies token-level thinking-in-speaking. SplitReason~\cite{splitreason-akhauri} offloads reasoning to backend LLMs. Qwen2.5-Omni~\cite{qwen2-tech-report-xu} presents a Thinker-Talker architecture in which the Talker takes high-level representations from the Thinker and generates speech tokens. Speculative and online speculative decoding~\cite{spec-decode-leviathan, online-spec-decoding-liu} allow faster generation that is later verified or refined by a larger model, improving effective decoding speed. \citet{predgen-li} extend this approach to cascaded speech models. \citeauthor{wavrag} facilitate retrieval from a text-audio hybrid knowledge
base. \citeauthor{streamrag} predict tool calls in parallel with user utterances. \citeauthor{kame} use an oracle-based LLM system to refine potential responses during user utterances. \citeauthor{moshirag} integrate different RAG backends with a full-duplex voice model frontend, handling retrieval delays of up to two seconds. 
\section{Conclusion}
\label{sec:conclusion}
We introduce conversational infill, a novel task leveraging a \talker-\reasoner approach. This enables lightweight and responsive SLM \talker models to respond promptly with context-aware fillers while also incorporating external knowledge from frontier-class \reasoner models into their responses at inference time. Consistently producing millisecond-level time-to-first-response across tasks with variable \reasoner{} latency and across \reasoner models, ConvFill maintains grounding in external knowledge while preserving task accuracy. ConvFill presents a promising approach to creating modular, responsive, intelligent voice agents for task-oriented dialogue.

Together, these properties place ConvFill at a new point on the latency--capability Pareto frontier~\citep{vera} by coupling the responsiveness of lightweight SLMs with the raw task performance of foundation models. We believe the architectural flexibility enabled by conversational infill presents a clear path toward developing conversational AI systems that excel at task-oriented and tool-use scenarios that are underserved by current state-of-the-art voice models. To encourage the broader community across AI and interactions research to experiment with and further validate the concepts presented in this paper, we release our model weights, dataset, and code for training and inference. 

\section*{Limitations}
\label{sec:limitations}
While most safety concerns in ConvFill's two-stage architecture are mitigated by a guardrailed \reasoner, the \talker model operates independently and may not immediately comply with safety constraints, as fillers are generated before \reasoner responses are available. Future work should explore tighter safety coupling between the two components. Our evaluation did not capture cases of \reasoner timeouts or excessive latency. In practice, prolonged filler sequences could become disruptive, and future work should investigate appropriate inter-filler pauses and fallback strategies for such scenarios. Additionally, a larger user evaluation sample would improve the statistical robustness and generalizability of our findings. Finally, future work could evaluate the applicability of ConvFill to languages other than English.
\section*{Ethical Considerations}
\label{sec:ethical-considerations}
Our user study involved human participants who were informed of the study's purpose and compensated for their participation. The study did not involve vulnerable populations and posed no foreseeable risk to participants. Our study was approved by the Institutional Review Board at our institution (Study \#{}STUDY00023284). Our synthetically generated dataset does not contain any personally identifiable information and was vetted for inappropriate content.

\section*{Acknowledgements}
\label{sec:acks}
We thank Maximus Powers for help with initial user interface prototyping and Alisa Liu for valuable feedback on early drafts of this paper. 

This research was supported in part by NSF awards 2338736 and 2310515. 

Vikram Iyer holds concurrent appointments as an Assistant Professor in the Paul G. Allen School of Computer Science and Engineering at the University of Washington and as an Amazon Visiting Academic. This paper describes work performed at the University of Washington and is not associated with Amazon.

\newpage
\bibliography{custom}
\clearpage
\onecolumn
\appendix
%
%
\lstdefinestyle{convfillprompt}{
    basicstyle=\footnotesize\ttfamily,
    breaklines=true,
    breakatwhitespace=false,
    columns=fullflexible,
    showstringspaces=false,
    keepspaces=true,
    language={},
    stringstyle=\color{black},
    keywordstyle=\color{black},
    commentstyle=\color{black},
    identifierstyle=\color{black},
}

\section{Appendix}
\label{appx:header}
This appendix contains additional details needed to reproduce the results of our experiments, as well as longer illustrative examples to support the main text. The appendix is laid out as follows:
\begin{itemize}
    \item \autoref{appx:dataset_generation}: Dataset Generation Details
    \item \autoref{appx:sec:model-training-details}: Model Training Details
    \item \autoref{appx:user-study}: User Study Details
    \item \autoref{appx:full-qa-results}: Full Benchmark Evaluation Results
    \item \autoref{appx:sec:inf-and-eval-prompts}: Inference and Evaluation Prompts
    
\end{itemize}

\subsection{AI Use Statement}
Large language model (LLM) assistants were used in the preparation of this work for programming assistance (including figure plotting and table generation), document formatting, and proofreading. All scientific content, experimental design, and conclusions are the work of the authors.

\newpage
\section{Dataset Generation}
\label{appx:dataset_generation}

The released dataset comes from two complementary generation tracks that
share a single scaffold-fill-validate-retry pipeline: a
\emph{topic-seeded freeform} track that synthesizes full conversations
from short topic prompts in six domains, and a \emph{DSTC8 schema-guided
scaffolded} track that re-skins existing task-oriented dialogues from
DSTC8~\cite{rastogi2020towards} into the conversational infill format while
pinning the original user utterances and system content as the thought
stream. Both tracks emit JSON conversations with the parallel
\texttt{thoughts}/\texttt{response} arrays shown in Fig.~\ref{fig:convfill-prompt-format} and pass through the same structural, NLI, and
BERTScore validators, with the scaffolded track adding two checks that
enforce preservation of the source content and the inference-time
visibility constraint. Validation thresholds throughout the pipeline
were selected by inspection. We describe
each track and the validators in turn; code, prompt templates, topic
lists, and the full configuration files will be released with the
dataset.

\subsection{Topic-Seeded Freeform Generation}
\label{appx:freeform}

\paragraph{Domain coverage and topic seeding.}
The freeform track covers six conversational domains: \emph{advice},
\emph{assistant queries}, \emph{event planning}, \emph{customer service},
\emph{education}, and \emph{medical conversation}. Each domain is keyed
by its own seed file of around 1,000 lines, with one conversation
generated per line. Two seeding strategies are used. For advice,
assistant queries, and event planning, each line is a free-text
description of the conversational topic (for example,
``\texttt{How can I get better at setting and following through on
goals?}''), paired with a fixed user persona such as
\texttt{a helpful assistant}. For customer service, education, and
medical conversations, each line is a \emph{$\langle$persona, scenario$\rangle$}
pair joined with a comma (``\texttt{Radiologic Technologist, The
technologist instructs the patient to hold their breath and remain
still for the chest X-ray.}''); the persona supplies the P2 role and
the scenario seeds the topic. We generated personas to be
domain-appropriate but not over-specified; highly specific personas
borrowed verbatim from PersonaHub~\cite{ge2025personahub} tended to
derail conversations toward unrelated biographical detail. Topic and
persona lists were generated up front with the same LLM used
for conversation generation and then frozen.

\paragraph{Per-conversation scaffold construction.}
For each topic, the generator samples the global shape of the
conversation, then renders a JSON skeleton with typed placeholders that
the LLM fills in. For each topic, we draw a turn count
$T \sim \mathcal{U}\{7,\ldots,10\}$, a per-turn silence schedule
$s_1,\ldots,s_T$ with $s_t \sim \mathcal{U}\{0,\ldots,3\}$ specifying
the number of contiguous \standout{<sil>} infill slots at the start of
each turn, and a per-turn substance count
$n_t \sim \mathcal{U}\{1,\ldots,5\}$ specifying the number of substance
lines after the infill block. Every position in the skeleton is a
typed placeholder (\texttt{<USER>}, \texttt{<INFILL\_N>},
\texttt{<THOUGHT\_N>}, \texttt{<RESPONSE\_N>}), and the
\texttt{thoughts} array already contains exactly $s_t$ leading
\texttt{"<sil>"} strings followed by \texttt{<THOUGHT\_N>} placeholders.
Pinning the silence schedule into the JSON structure makes it a
property of the prompt rather than something the LLM must obey through
natural-language instructions alone. A representative turn skeleton
for $s_t=2, n_t=3$ is:

\begin{tcolorbox}[colback=gray!3,colframe=gray!40,boxsep=2pt,left=4pt,right=4pt,top=2pt,bottom=2pt]
\footnotesize\ttfamily
\{\\
~~"user": "<USER>",\\
~~"thoughts": ["<sil>", "<sil>", "<THOUGHT\_1>", "<THOUGHT\_2>", "<THOUGHT\_3>"],\\
~~"response": ["<INFILL\_1>", "<INFILL\_2>", "<RESPONSE\_1>", "<RESPONSE\_2>", "<RESPONSE\_3>"]\\
\}
\end{tcolorbox}

The skeleton is wrapped in a \texttt{conversation} array of $T$ such
turns and serialized as JSON.

\paragraph{Prompt assembly.}
We use a three-message chat structure: a user message containing the
rendered instruction template and one in-context example conversation,
a fixed assistant acknowledgment that the model will preserve the
scaffold structure, and a final user message containing the JSON
skeleton to be filled. The fixed acknowledgment pushes the model past
its planning step and makes the skeleton the most recent input it
conditions on. The freeform prompt template is shown in
Prompt~\ref{prompt:freeform}. The \texttt{\{example\}} slot at the
bottom is filled with a single hand-authored example conversation per
domain, which anchors the style of \emph{thoughts} (terse, single-idea,
factual) and \emph{responses} (conversational, paired one-to-one with
the corresponding thought). NOTE: the prompt has been modified slightly
to fit within the document. To precisely re-create our setup, use the
full prompt files from the project repository. 

\begin{prompt}[title={Prompt \thetcbcounter: Freeform conversation generation template}, label=prompt:freeform, breakable]
\begin{lstlisting}[style=convfillprompt]
Generate a {num_turns}-turn conversation with two people, P1 and P2.
    * All conversations are spoken out loud, so spell out any numbers or
      expressions and remove quotation marks around words.
    * Both sides should be proactive and contribute to the conversation.
    * P1 is {P1}
    * P2 is {P2}
    * Conversation topic: {topic}
    * Conversation MUST be EXACTLY {num_turns} TURNS LONG and end naturally.
    * All turns MUST have content.
    * Allowed characters in user, thoughts, and response strings: English
      letters, digits, spaces, and these punctuation marks ONLY:
      . , ! ? ; : ' - / ( ) "
      Spell out anything else (e.g. "and" not "&", "dollars" not "$").
    * Do not include names; all references should be in first or second
      person.

Response Rules (how P2 actually speaks):
    * The response array represents P2 speaking in real time, line by line.
    * Some lines are "infill": contextually appropriate dialogue P2 says
      BEFORE having the full answer. These are NOT generic fillers like
      "hmm" or "let me think." Instead, they are natural continuations of
      the conversation: acknowledging what P1 said, restating part of the
      question, offering a relevant opinion, showing understanding, and so
      on.
    * The remaining lines deliver substantive information conversationally.
    * Each response entry is the conversational delivery of its
      corresponding thought. Keep responses tight rather than
      paragraph-length.
    * VARY the number of sentences per turn; every one should be different.
    * Turns should be standalone (do not refer to previous turns).

INFILL LINE SPECIFICATION:
    Each turn has a specified number of infill lines.
    Good infill lines:
      - Acknowledge or react to what P1 specifically said
      - Restate or reflect on the question in P2's own words
      - Share a brief relevant opinion that connects to the topic
      - Show genuine engagement with the specific situation
    Bad infill lines (do NOT produce):
      - Generic fillers ("Hmm", "Let me think", "Good question")
      - Empty acknowledgments that could apply to any conversation
      - Phrases that stall without adding contextual value

    Per-turn infill counts (MUST be followed exactly):
{sil_schedule_text}

    * If a turn specifies 0 infill lines, P2 jumps straight to substance.
    * If a turn specifies N infill lines, the first N entries in response
      must be infill, and the first N entries in thoughts must be "<sil>".
    * After the infill lines, each turn MUST have between {substance_min}
      and {substance_max} substantive lines.

INFILL QUALITY REQUIREMENTS:
    * Every infill line must be SPECIFIC to what P1 just said.
    * Do NOT reuse the same phrase or pattern across turns.
    * Infill lines should vary naturally and flow into the substantive
      content that follows.

Output in JSON Format (ONLY the JSON, no commentary):
    * Wrap output in a "conversation" array containing turn objects.
    * Each turn object must have:
        * user:     What P1 says
        * thoughts: Concise, direct information for each part of P2's turn
        * response: How P2 actually says it in conversation
    * ALL FIELDS MUST BE NON-EMPTY.
    * thoughts and response arrays MUST have the same number of elements.

Thoughts rules (the core information P2 wants to convey):
    * Infill entries MUST be exactly "<sil>".
    * Non-infill thoughts should read like a knowledgeable person's
      direct, concise answer. No small talk, no personality, no filler;
      just the core information.
    * SAME VOICE AS RESPONSE. The thought IS the content P2 conveys, not
      a self-instruction or third-person description.
    * ONE IDEA PER THOUGHT. NO compound sentences linked by "because",
      "while", "and", "which", "since", "so", or semicolons that combine
      distinct ideas.
    * If you have more ideas than thought slots, pick the most important
      and drop the rest.
    * Each thought must capture the full meaning of its corresponding
      response entry, in order.

You will be given a JSON scaffold to complete. Replace each placeholder:
- <USER>:       What P1 says
- <THOUGHT_N>:  Concise, direct, factual thought
- <RESPONSE_N>: Conversational delivery of the corresponding thought
- <INFILL_N>:   Contextual dialogue continuing the conversation
- "<sil>" entries must remain exactly as "<sil>"

Do NOT add, remove, or reorder any entries. Return ONLY the completed
JSON.

Example conversation:
{example}
\end{lstlisting}
\end{prompt}

\subsection{DSTC8 Schema-Guided Scaffolded Generation}
\label{appx:scaffold}

The freeform track produces stylistically homogeneous, opinion- and
advice-heavy dialogues but underrepresents goal-directed slot-filling
exchanges (bookings, queries, and transactional confirmations). To broaden
coverage, we add a second track that takes existing task-oriented
dialogues from DSTC8~\cite{rastogi2020towards} and re-skins them into the
conversational infill format, keeping the original user and system
content while letting the LLM author the infill and response delivery.

\paragraph{Source corpus.}
We sample from the DSTC8 \texttt{train} split, which contains 26
services spanning roughly 16 domains (banks, buses, calendar, events,
flights, homes, hotels, media, movies, music, rental cars, restaurants,
ridesharing, services, travel, weather, and others). Each dialogue is a
sequence of strictly alternating USER and SYSTEM turns with verbatim
utterances and service annotations.

\paragraph{DSTC8-to-scaffold conversion.}
Each dialogue is converted to a conversation scaffold by walking through turns
in order and pairing every USER turn with the following SYSTEM turn:

\begin{tightitemize}
\item the USER utterance is pinned verbatim as the turn's \texttt{user}
string;
\item the SYSTEM utterance is sentence-segmented using a regex split on
sentence-final punctuation; the resulting sentences become the
substance entries of the \texttt{thoughts} array;
\item a turn-specific silence count $s_t \sim \mathcal{U}\{0,\ldots,3\}$
is prepended as \texttt{"<sil>"} entries, matching the freeform
convention;
\item the \texttt{response} positions are emitted as \texttt{<INFILL\_N>}
and \texttt{<RESPONSE\_N>} placeholders to be filled by the LLM.
\end{tightitemize}

Unlike the freeform track, the LLM in this track fills only the
\texttt{response} array; \texttt{user} and \texttt{thoughts} entries are
pre-filled with real DSTC8 content and must be reproduced verbatim
(enforced by structural validation, \S\ref{appx:validation}).

\paragraph{Inference-time visibility constraint.}
Because each scaffold encodes a complete DSTC8 dialogue, the LLM sees
the entire conversational arc, including all future user utterances and
all future SYSTEM content, while authoring infills for the earliest
turns. This creates a causality mismatch between generation and
inference: at deployment, the ConvFill \talker only sees turns that have already
been spoken (the previous turn and the current turn so far), but the
generating LLM has access to the whole future of the dialogue and tends
to leak details from later turns into earlier infills. A common failure
is producing an infill that references a proper noun (a restaurant
name, a movie title, or an address) that has not yet been mentioned at the
point in the dialogue where the infill is supposed to be spoken.

We address this in two places. First, the scaffold prompt template
defines an explicit \emph{visibility window}: when generating an infill
at turn $t$, the model is told to ground it only in the previous turn
(the user message and the assistant response) and the current turn so
far (the user message and any earlier infills already placed). Second,
Stage~4 of the validation cascade enforces the constraint
programmatically using a proper-noun visibility check
(\S\ref{appx:validation}), giving an extra layer of robustness against
leaks that slip past the prompt.

The scaffold prompt template is largely identical to the freeform one
in Prompt~\ref{prompt:freeform}; it removes the topic and turn-count
slots (both implicit in the scaffold), replaces the freeform thought
rules with the response rules below, and adds the visibility-window
constraint. The substantive changes are shown in
Prompt~\ref{prompt:scaffold-deltas}.

\begin{prompt}[title={Prompt \thetcbcounter: DSTC8 scaffold deltas (substantive additions to Prompt~\ref{prompt:freeform})}, label=prompt:scaffold-deltas]
\begin{lstlisting}[style=convfillprompt]
You are completing a partial conversation between two people, P1 and P2.
The "user" turns and the "thoughts" arrays are already filled in by an
external source; you only fill the response arrays.
    * P1 is {p1}
    * P2 is {p2}
    * The conversation comes from a {service_pretty} domain; P2 is a
      knowledgeable assistant for that domain.

VISIBILITY WINDOW:
    Infill visibility: when P2 produces an infill line, the only visible
    context is the previous turn (P1's user message and P2's complete
    response) and the current turn so far (P1's user message and any
    earlier infill lines already produced in this turn's response array).
    Each infill must be grounded only in that visible context.

    Concrete example of the most common failure mode:
    Suppose in turn 1 the user asked to play "Casablanca" and the
    assistant queued it. Many turns later, the dialogue ends:

      previous turn: user "Yes, play now."
                     response "Great, your movie is starting now! Enjoy!"
      current turn:  user "Thank you. Good bye."
                     thoughts ["<sil>", "<sil>", "<sil>", "you're welcome."]

      WRONG infill: "Hope you enjoy Casablanca."
        ^ "Casablanca" was named MANY turns ago, not in the previous turn
          or the current turn. Forbidden, even though it sounds natural.
      RIGHT infill: "It was my pleasure helping you out today."
        ^ grounded in the immediate goodbye exchange.

RESPONSE-ONLY FILLING:
    * user strings MUST be copied verbatim. Any change is a failure.
    * thoughts non-"<sil>" entries MUST be copied verbatim. Any change
      is a failure.
    * "<sil>" entries in thoughts MUST remain exactly as "<sil>".
    * Each response must capture the FULL meaning of the corresponding
      thought and INTRODUCE NO NEW INFORMATION beyond what the paired
      thought contains. No extra facts, no new offers, no new follow-up
      questions, no extra details.
\end{lstlisting}
\end{prompt}

\subsection{Validation Cascade}
\label{appx:validation}

Every generated conversation passes through a four-stage cascade.
Stages~1--3 apply to both tracks; Stage~4 applies only to the DSTC8
scaffolded track, where the source content gives the validator extra
ground truth. A conversation must pass every stage to be admitted; a
failure at Stage~1 triggers a single in-chat correction request, and a
failure at any stage that survives correction counts against the
per-topic retry budget. Threshold values reported below were chosen by inspection.

\paragraph{Stage 1: Structural validation.}
Checks performed on each turn and the conversation as a whole are
summarized in Table~\ref{tab:structural-checks}. Key rules:
\standout{<sil>} entries must be contiguous at the start of the
\texttt{thoughts} array; the per-turn \standout{<sil>} count must match
the scaffold; \texttt{thoughts} and \texttt{response} arrays must have
equal length per turn; no infill phrase (case-folded and trimmed) may
be reused more than two times across the entire conversation
(\texttt{MAX\_FILLER\_REUSE}\,=\,2); and generated text uses a
restricted spoken character set (English letters, digits, whitespace,
and the punctuation \texttt{.\,,\,!\,?\,;\,:\,'\,-\,/\,(\,)\,"}).

\begin{table}[t]
\centering
\small
\begin{tabular}{p{0.34\linewidth}p{0.58\linewidth}}
\toprule
\textbf{Check} & \textbf{Failure condition} \\
\midrule
JSON parse        & Output is not valid JSON. \\
Placeholder leak  & \texttt{<USER>}, \texttt{<THOUGHT\_N>}, etc.\ remain in the output. \\
Required fields   & Any turn missing \texttt{user}, \texttt{thoughts}, or \texttt{response}. \\
Turn count        & $T \notin [7,10]$ (freeform only; scaffold uses DSTC8 length). \\
Equal-length arrays & $|\text{thoughts}| \neq |\text{response}|$ in any turn. \\
Sil count         & $\#\{\texttt{<sil>}\} \neq s_t$ in turn $t$. \\
Sil contiguity    & \standout{<sil>} not in positions $0,\ldots,s_t{-}1$, or any \standout{<sil>} at position $\geq s_t$. \\
Substance count   & $n_t \notin [1,5]$ (freeform) or $[1,10]$ (scaffold). \\
Min substance length & Any non-\standout{<sil>} thought shorter than 5 characters (10 in scaffold). \\
Character set     & Any disallowed symbol in \texttt{response} (and in \texttt{user}/\texttt{thoughts} for freeform; relaxed for scaffold since DSTC8 source can contain \$, \%, etc.). \\
Filler diversity  & Any infill phrase used more than 2$\times$ in the conversation. \\
Scaffold pin      & (scaffold only) \texttt{user} or non-\standout{<sil>} \texttt{thoughts} not byte-equal to source. \\
\bottomrule
\end{tabular}
\caption{Stage 1 structural validation checks.}
\label{tab:structural-checks}
\end{table}

On a structural failure, the pipeline issues one in-chat correction
request: the model sees its own bad output and a targeted error
message, and is asked to return a corrected JSON in the same chat. If
the correction also fails, the topic counts as one attempt and is
retried from scratch with a fresh schedule.

\paragraph{Stage 2: NLI semantic validation.}
We score every non-\standout{<sil>} (thought, response) pair using a
publicly available DeBERTaV3-base model fine-tuned on
MNLI~\cite{he_debertav3,williams_broad-coverage_2018}\footnote{The
\texttt{MoritzLaurer/DeBERTa-v3-base-mnli} checkpoint from the
Hugging Face Hub.}, treating the thought as the premise and the response
as the hypothesis. Two gates are applied:

\begin{tightitemize}
\item \textbf{Per-pair gate.} For each non-\standout{<sil>} pair
$(t_i, r_i)$ in any turn, reject the conversation if the contradiction
probability $P(\text{contradiction} \mid t_i, r_i) > 0.20$.
\item \textbf{Whole-turn gate.} For each turn, the premise is the user
utterance concatenated with all non-\standout{<sil>} thoughts, and the
hypothesis is the user utterance concatenated with all response
entries. Reject if the whole-turn contradiction probability exceeds
$0.30$.
\end{tightitemize}

The per-pair gate catches local divergences where a single response
changes the meaning of its paired thought. The whole-turn gate catches
aggregate drift where each pair is individually plausible but the turn
as a whole reframes the underlying message.

\paragraph{Stage 3: BERTScore alignment validation.}
NLI catches semantic contradiction within a (thought, response) pair
but not \emph{positional misalignment}: a response at position $i$
whose content semantically belongs to position $j$ in the same turn.
We catch this with BERTScore~\cite{zhang2020bertscore} F1 using
\texttt{microsoft/deberta-xlarge-mnli} (layer 40), applying two
constraints at every non-\standout{<sil>} position $i$:

\begin{tightitemize}
\item \textbf{Anchor.} $\text{F1}(t_i, r_i) \geq 0.70$: the response at
position $i$ must be at least moderately similar to the thought at
position $i$.
\item \textbf{Cross.}
$\text{F1}(t_i, r_i) > \max_{j \neq i} \text{F1}(t_i, r_j)$: the
response at position $i$ must be \emph{more} similar to its paired
thought than any other response in the same turn (infill positions
included). If some other response is more aligned with $t_i$, the
arrays are likely transposed or duplicated, and we reject.
\end{tightitemize}

\paragraph{Stage 4: Scaffold-preservation checks (DSTC8 only).}
Two additional checks run for the scaffolded track.

\begin{tightitemize}
\item \textbf{Thought consistency.} For each turn, we compute BERTScore
F1 between the joined non-\standout{<sil>} thoughts and the verbatim
source SYSTEM utterance from DSTC8. Reject if $\text{F1} < 0.95$. The
\texttt{thoughts} array is already string-pinned by Stage~1, so F1
should be near $1.0$; this floor is a backstop against drift introduced
by sentence-splitting edge cases.
\item \textbf{Proper-noun visibility window.} This is the programmatic
implementation of the visibility constraint motivated in
\S\ref{appx:scaffold}. We extract Title-Case mid-sentence proper nouns
from each response line using a regex
(\emph{Casablanca}, \emph{San Francisco}, \emph{Auzerais Avenue}). For
each detected proper noun in $\text{response}_{t,i}$, we check
case-insensitive membership in the union of (a) the current turn's
user utterance, (b) the current paired thought $\text{thoughts}_{t,i}$,
(c) earlier non-\standout{<sil>} thoughts in turn $t$, (d) the previous
turn's user utterance, and (e) the previous turn's non-\standout{<sil>}
thoughts. Previous-turn responses are intentionally excluded from the
allowed corpus: they are LLM-generated and may themselves carry leaked
content. Any proper noun outside this window causes the conversation
to be rejected.
\end{tightitemize}

\subsection{Generation Settings}
\label{appx:hyperparams}

Table~\ref{tab:hyperparams} reports the hyperparameters needed to
reproduce the dataset. The full per-domain configuration JSON files
will be released with the dataset.

\begin{table}[t]
\centering
\footnotesize
\begin{tabular}{p{0.50\linewidth}p{0.42\linewidth}}
\toprule
\textbf{Setting} & \textbf{Value} \\
\midrule
Generation LLM             & Claude Opus 4.6 \\
Decoding temperature       & 1.0 \\
Max output tokens          & 4000 \\
\midrule
Turn count $T$ (freeform)  & $\mathcal{U}\{7,10\}$ \\
Turn count $T$ (scaffold)  & DSTC8 dialogue length \\
Per-turn sil count $s_t$   & $\mathcal{U}\{0,3\}$ \\
Per-turn substance $n_t$ (freeform) & $\mathcal{U}\{1,5\}$ \\
Per-turn substance $n_t$ (scaffold) & DSTC8 sentence count \\
\midrule
Min substance length       & 5 chars (freeform), 1 (scaffold) \\
Max infill phrase reuse    & 2 \\
\midrule
NLI model                  & DeBERTaV3-base-MNLI \\
Per-pair contradiction max & 0.20 \\
Whole-turn contradiction max & 0.30 \\
\midrule
BERTScore model            & DeBERTa-xlarge-MNLI, layer 40 \\
Anchor F1 floor            & 0.70 \\
Scaffold thought F1 floor  & 0.95 \\
\bottomrule
\end{tabular}
\caption{Generation and validation hyperparameters.}
\label{tab:hyperparams}
\end{table}

\subsection{Output Format and Yield}
\label{appx:stats}

Each accepted conversation is emitted as a single JSON object on its
own line of a JSONL file, with the shape
\texttt{\{"conversation": [\{"user", "thoughts", "response"\}, ...]\}}.
Scaffolded conversations additionally carry a top-level
\texttt{scaffold\_metadata} object recording the DSTC8 service and
dialogue ID. 

Table~\ref{tab:yield} reports the per-track yield. The released dataset
contains 6{,}005 freeform conversations (50{,}324 turns) and
2{,}438 DSTC8-scaffolded conversations (24{,}184 turns), for a total of
8{,}443 conversations and 74{,}508 turns.

\begin{table}[t]
\centering

\begin{tabular}{lrrr}
\toprule
\textbf{Track / Domain} & \textbf{Convs.} & \textbf{Turns} & \textbf{Turns/Conv.} \\
\midrule
Advice                & 1{,}000  & 8{,}308  & 8.31 \\
Assistant queries     & 1{,}000  & 8{,}408  & 8.41 \\
Customer service      & 1{,}000  & 8{,}313  & 8.31 \\
Education             & 1{,}000  & 8{,}421  & 8.42 \\
Medical               & 1{,}000  & 8{,}400  & 8.40 \\
Event planning        & 1{,}005  & 8{,}474  & 8.43 \\
\midrule
Freeform subtotal     & 6{,}005  & 50{,}324 & 8.38 \\
DSTC8 scaffolded      & 2{,}438  & 24{,}184 & 9.92 \\
\midrule
\textbf{Total}        & \textbf{8{,}443} & \textbf{74{,}508} & 8.83 \\
\bottomrule
\end{tabular}
\caption{Per-track conversation and user-infill turn counts in the
released dataset.}
\label{tab:yield}
\end{table}
\newpage
\section{Model Training Details}
\label{appx:sec:model-training-details}
This section contains details about the model training pipeline and model training parameters used in our experiments. It is intended to provide an overview of how to reproduce or extend the experiments in this paper.

\subsection{Base SLM Models}
\label{appx:subsec:talker-base-models}
The table below contains details about the Base SLM models that were fine-tuned to perform the conversational infill task, including the Hugging Face ID of each model.

\begin{table}[h]
\centering
\small
\begin{tabular}{lcccp{5.5cm}}
\toprule
\textbf{Model Family} & \textbf{\# Param.} & \textbf{Size (BF16)} & \textbf{Size (INT8)} & \textbf{Hugging Face ID} \\
\midrule
Gemma3 IT   & 270M  & 511 MB   & 272 MB   & \texttt{google/gemma-3-270m-it} \\
Gemma3 IT   & 1B    & 1.86 GB  & 1013 MB  & \texttt{google/gemma-3-1b-it} \\
Qwen3       & 0.6B  & 1.40 GB  & 604 MB   & \texttt{Qwen/Qwen3-0.6B} \\
SmolLM2 IT  & 135M  & 257 MB   & 136 MB   & \texttt{HuggingFaceTB/SmolLM2-135M-Instruct} \\
SmolLM2 IT  & 360M  & 690 MB   & 367 MB   & \texttt{HuggingFaceTB/SmolLM2-360M-Instruct} \\
SmolLM2 IT  & 1.7B  & 3.19 GB  & 1.69 GB  & \texttt{HuggingFaceTB/SmolLM2-1.7B-Instruct} \\
Llama 3.2 IT & 1.2B  & 2.30 GB  & 1.22 GB  & \texttt{meta-llama/Llama-3.2-1B-Instruct} \\
\bottomrule
\end{tabular}
\caption{\textbf{Fine-tuned Base SLM models.} Seven models were fine-tuned to perform the conversational infill task. ``IT'' indicates that the base model was originally instruction-tuned.}
\label{tab:convfill-talkers-finetuned}
\end{table}

\subsection{Frontier Models}
\begin{table}[h]
\centering
\small
\begin{tabular}{llll}
\toprule
\textbf{Model Name} & \textbf{Accessed Through} & \textbf{Model Identifier} & \textbf{Role(s)} \\
\midrule
GPT-5.5   & OpenAI API  & \texttt{gpt-5.5-2026-04-23} & ConvFill \reasoner\\
Claude Opus 4.7   & Anthropic API  & \texttt{claude-opus-4-7} & ConvFill \reasoner\\
Gemini 3.1 Pro  & Google AI Studio  & \texttt{3.1-pro-preview-01-2026} & ConvFill \reasoner\\
GPT-4o & OpenAI API  & \texttt{gpt-4o-2024-08-06} & Automated Grading\\
Claude Opus 4.6   & Anthropic API  & \texttt{claude-opus-4-6} & Dataset Generation\\
\bottomrule
\end{tabular}
\caption{\textbf{API-based models used for development, inference, and testing.} Model identifiers denote the specific tagged versions used in this work.}
\label{tab:convfill-backend-apis}
\end{table}
 
\subsection{Control Sequence Expansions for Different \talker Models}
\label{appx:subsec:ctrl-tokens}

We use existing model control tokens as boundary tokens to mark spans for user, assistant, and knowledge segments of training and inference examples.

\begin{table*}[h]
\small
\centering
\begin{tabular}{lll}
\toprule
\textbf{Model} & \textbf{Control Token Sequence} & \textbf{Expansion} \\
\midrule
\multirow{4}{*}{Gemma 3}
    & \standout{[USER]}       & \texttt{<start\_of\_turn>user} \\
    & \standout{[KNOWLEDGE]}  & \texttt{<start\_of\_turn>knowledge} \\
    & \standout{[ASST]}       & \texttt{<start\_of\_turn>model} \\
    & \standout{[END]}        & \texttt{<end\_of\_turn>} \\
\midrule
\multirow{4}{*}{Llama 3.2}
    & \standout{[USER]}       & \texttt{<|start\_header\_id|>user<|end\_header\_id|>} \\
    & \standout{[KNOWLEDGE]}  & \texttt{<|start\_header\_id|>knowledge<|end\_header\_id|>} \\
    & \standout{[ASST]}       & \texttt{<|start\_header\_id|>assistant<|end\_header\_id|>} \\
    & \standout{[END]}        & \texttt{<|eot\_id|>} \\
\midrule
\multirow{4}{*}{Qwen3}
    & \standout{[USER]}       & \texttt{<|im\_start|>user} \\
    & \standout{[KNOWLEDGE]}  & \texttt{<|im\_start|>knowledge} \\
    & \standout{[ASST]}       & \texttt{<|im\_start|>assistant} \\
    & \standout{[END]}        & \texttt{<|im\_end|>} \\
\midrule
\multirow{4}{*}{SmolLM2}
    & \standout{[USER]}       & \texttt{<|im\_start|>user} \\
    & \standout{[KNOWLEDGE]}  & \texttt{<|im\_start|>knowledge} \\
    & \standout{[ASST]}       & \texttt{<|im\_start|>assistant} \\
    & \standout{[END]}        & \texttt{<|im\_end|>} \\
\bottomrule
\end{tabular}
\caption{Control sequence expansions per model. \standout{[KNOWLEDGE]} is a new role introduced by our pipeline and mapped through each model's existing control-token template.}
\label{tab:boundary_tokens}
\end{table*}

\subsection{Shared Training Parameters for \texorpdfstring{\talker}{Talker} Models}
\label{appx:subsec:shared-training-params}

The following describes the \talker model training parameters. Each \talker model was trained on a 90/10 train/evaluation split made up of the dataset in \autoref{appx:dataset_generation}. Models were trained on either 4 NVIDIA RTX 6000 GPUs or 4 NVIDIA A40 GPUs. GPU hours are documented in \autoref{tab:talker-training-configs}.

\begin{table}[h]
\centering
\small
\begin{tabular}{llccccc}
\toprule
\textbf{Model} & \textbf{Parameters} & \textbf{Batch Size} & \textbf{Eff.\ Batch Size} & \textbf{Chat Template} & \textbf{GPU Hours} & \textbf{Cost (USD)} \\
\midrule
SmolLM2 IT   & 135M  & 4 & 16 & ChatML  & 2.80  & \$3.07  \\
Gemma 3 IT   & 270M  & 4 & 16 & Gemma   & 4.53  & \$4.96  \\
SmolLM2 IT   & 360M  & 4 & 16 & ChatML  & 11.60 & \$12.70 \\
Qwen3        & 0.6B  & 4 & 16 & ChatML  & 6.94  & \$7.60  \\
Gemma 3 IT   & 1B    & 2 & 8  & Gemma   & 31.70 & \$34.71 \\
Llama 3.2 IT & 1B    & 2 & 8  & Llama-3 & 15.60 & \$17.08 \\
SmolLM2 IT   & 1.7B  & 2 & 8  & ChatML  & 48.93 & \$53.58 \\
\midrule
\textbf{Total} & & & & & \textbf{122.10} & \textbf{\$133.70} \\
\midrule
\multicolumn{7}{l}{\textit{Shared hyperparameters (all models)}} \\
\midrule
\multicolumn{2}{l}{Learning rate}         & \multicolumn{5}{l}{$1 \times 10^{-4}$} \\
\multicolumn{2}{l}{LR schedule}           & \multicolumn{5}{l}{Cosine with warmup} \\
\multicolumn{2}{l}{Warmup steps}          & \multicolumn{5}{l}{200} \\
\multicolumn{2}{l}{Weight decay}          & \multicolumn{5}{l}{0.01} \\
\multicolumn{2}{l}{Gradient accumulation} & \multicolumn{5}{l}{4} \\
\multicolumn{2}{l}{Epochs}                & \multicolumn{5}{l}{1} \\
\multicolumn{2}{l}{Special tokens}        & \multicolumn{5}{l}{\standout{<sil>}} \\
\bottomrule
\end{tabular}
\caption{Training configurations for all fine-tuned \talker models. Effective batch size = batch size $\times$ gradient accumulation steps. GPU hours = wall-clock time $\times$ 4 GPUs. Cost estimated at \$1.095/GPU/hr~\citep{googlecloud2026gpupricing}.}
\label{tab:talker-training-configs}
\end{table}
\newpage

\section{User Study Details}
\label{appx:user-study}
This section provides an in-depth description of our user study procedures for the conversational infill study.

\subsection{Ethical Considerations} 
\label{appx:user-study-ethical-considerations}
Our study was approved by the Institutional Review Board at our institution. Participants engaged in sessions of approximately 45 minutes for both the interaction and prompt grading portions of the study. Participants were informed that their participation in our study was voluntary and that the data generated during the study would be included in an academic publication. Participants were compensated for their participation.

\subsection{Study Instructions} 
\label{appx:user-study-instructions}
Participants were provided with the following prompt to begin the study. \\

\textit{You will interact with three different systems. After interacting with each system, you will fill out a survey with some questions about the experience. There will be three blocks in the study where you will perform three different tasks. At the end of the block, you will rank each of the systems in order of preference.}

\textit{You will be provided with a sheet to guide your conversation. The aim of this sheet is to give you a goal to accomplish and suggest things to say. For each of the tasks, use the system to gather the information and insights you desire. Ask whatever questions feel natural to you; there is no script to follow. Work through the task in whatever order makes sense. Converse with the system for several turns. Once you have formed an opinion about it (e.g., after 4--5 turns of conversation), you may stop.}

\textit{As you interact with each agent, pay attention to how you feel about the interaction. Notice things like:
\begin{enumerate}[noitemsep]
    \item How easy or difficult it is to get the information you need
    \item How the agents respond
    \item How helpful the agent's responses are
    \item How the conversation flows
    \item Anything that you like or dislike about the interaction
\end{enumerate}}

After the introductory prompt, participants were introduced to the metrics that they would use for Likert-scale ratings later in the study. The following descriptions were provided, without labels. However, we provide the labels here for clarity and so that they can be referenced with the body of the text:
\textit{
\begin{enumerate}[noitemsep]
    \item Latency: The system responded promptly enough to maintain a natural conversation.
    \item Clarity: I could understand the words the system said. 
    \item Fluency: I could understand the ideas expressed by the system.
    \item Response Length: The length of the responses felt appropriate. 
    \item Coherence: Sentences related to each other and progressed in a logical fashion.
    \item Task Completion: The system's responses provided the information I needed to complete the assigned task.
    \item Naturalness: The language used felt natural and appropriate for the context.
    \item Satisfaction: I was satisfied with my conversation(s) with the system. 
\end{enumerate}}

\subsection{Study Tasks} 
\label{appx:subsec:study-tasks}
The following section contains details about the tasks that were assigned to participants in the categories of Direct, RAG, and MCP.

\subsubsection{Direct Tasks}
\label{appx:subsubsec:normal-tasks}
Direct tasks refer to task-oriented dialogues that can progress without necessarily having access to tools, external databases, or external tools. Examples include trip planning, looking up recipes, brainstorming ideas for writing, and planning events. Three different scenarios of this style were presented to participants, and they were instructed to choose the one that they wanted to have a conversation about. Locations and names are redacted to preserve anonymity during the submission process. All three system configurations (Base SLM, Frontier, and ConvFill) are used for this task.\\

\noindent
\textbf{Scenario 1: Birthday Party Planning} Your task is to plan a birthday party for your friend using the agent. You need to figure out where to host it, what activities to include, the costs involved, how to decorate, and any other details that matter to you. Talk to the agent naturally and ask whatever questions you need. There is no single right way to do this; explore the information in whatever order makes sense to you, and feel free to change your mind as you learn more. Suggestions for conversation:
\begin{itemize}[noitemsep]
  \item Venue(s) for hosting 20 people for a birthday party in [LOCATION]
  \item Primary activity or activities for the party (you can choose)
  \item Cost estimates for activities (your budget is decided by you)
  \item Vendor policies (e.g., can they accommodate outside food/drinks at any of the venues?)
  \item Decoration ideas that match the theme (you choose the theme)
  \item Timeline/logistics for setup (e.g., how far in advance to order some supplies)
  \item Any additional details you discovered or decided on
\end{itemize}

\noindent
\textbf{Scenario 2: Train Travel} Your task is to plan a train trip from [LOCATION A] to [LOCATION B] using the agent. You need to figure out which route(s) make sense, how long the trip takes, what it costs, what discounts you might qualify for, and what to expect in terms of food and amenities on the train. Talk to the agent naturally and ask whatever questions you need. There is no single right way to do this; explore the information in whatever order makes sense to you. Suggestions for conversation:
\begin{itemize}[noitemsep]
  \item Available train route options from [LOCATION A] to [LOCATION B]
  \item Whether direct routes exist or if connections are required
  \item Total travel time for your chosen route(s)
  \item Approximate ticket price(s)
  \item Student discount options and how to apply them
  \item Food options available on the train
  \item Onboard amenities (sleeping accommodations, showers, etc.)
  \item Any other details you discovered or questions you resolved
\end{itemize}

\noindent
\textbf{Scenario 3: Baking a Cake} Your task is to use the agent to find a recipe for making a cake that uses Greek yogurt. You need to figure out what recipe to use, how to incorporate the texture, ingredients, and toppings you want, how long it will take to prepare and bake, and whether you can realistically finish it before your friends arrive. Talk to the agent naturally and ask whatever questions you need. There is no single right way to do this; explore the information in whatever order makes sense to you. Suggestions for conversation:
\begin{itemize}[noitemsep]
  \item Recipe for a Greek yogurt cake
  \item How to make the cake fluffy
  \item How to incorporate other ingredients that you like (you can choose)
  \item Some sort of topping that will go with the cake
  \item Total time needed for preparation and baking
  \item Whether the recipe fits your timeline before your friends arrive
  \item Any ingredient substitutions or variations you want
\end{itemize}

\subsubsection{RAG Task}
\label{appx:subsubsec:rag-task}
This task connects the conversational model to a retrieval-augmented generation (RAG) database. All three system configurations (Base SLM, Frontier, and ConvFill) are used for this task. The RAG database was constructed by scraping a set of webpages related to a certain institution. The RAG system operates as a two-stage retrieve-then-rerank pipeline over a pre-built index of [INSTITUTION]'s documentation. At inference time, the user query is embedded using OpenAI \texttt{text-embedding-3-large} (3,072 dimensions) and L2-normalized before being issued to a FAISS index containing 367 pre-embedded document chunks. The top 30 candidate chunks are retrieved in this first stage. The 30 candidates are then passed to a cross-encoder reranker, which scores each (query, chunk) pair jointly and reorders the candidates by relevance. The top three reranked chunks are passed to the downstream language model for grounded response generation.\\

\noindent
The participant prompt for the RAG task is as follows:\par
Your friend is a [PERSON] at the [INSTITUTION] and they are just getting started. They gave you a list of some questions that they want to find out about, and your job is to answer those questions. You do not need to specify the institution; simply ask the questions. Your task is to help your friend find answers to their questions. There is no single right way to do this; explore the information in whatever order makes sense to you, and feel free to change your mind as you learn more.

Just like above, a list of suggested topics is provided. However, to maintain anonymity during the submission period, this list is excluded.

\subsubsection{MCP Task}
\label{appx:subsubsec:mcp-task}
In the MCP task, models are connected via Model Context Protocol to an IMAP email server. In this task, we support only the Frontier and ConvFill system configurations because Base SLM models are unable to make properly formatted MCP calls.

The participant prompt for the MCP task is as follows:
You are a busy person, and you need to check some things related to your email. You have a conversational agent that can help you out. Use the conversational agent to ask questions about your email. The agent can see your inbox and will respond appropriately.

Your task is to answer some of the following questions on your checklist. There is no single right way to do this; explore the information in whatever order makes sense to you, and feel free to change your mind as you learn more. Suggestions for conversation:
\begin{itemize}[noitemsep]
  \item Number of emails in inbox: \underline{\hspace{3cm}}
  \item Any unread emails: \underline{\hspace{3cm}}
  \item \texttt{[EVENT]} date: \underline{\hspace{3cm}}
  \item Summary of last on-call handoff: \underline{\hspace{3cm}}
  \item Any build failures: \underline{\hspace{3cm}}
  \item Rust updates for the week: \underline{\hspace{3cm}}
  \item Deadline for Q1 roadmap review: \underline{\hspace{3cm}}
\end{itemize}

\subsection{Participant Condition Assignment} 
\label{appx:subsec:condition-assignment}
The eighteen participants were assigned to three experimental conditions (tasks: Direct, RAG, MCP) in a fully counterbalanced Latin square design, such that each condition appeared exactly six times in each ordinal position (first, second, third) across participants. Within each condition, participants evaluated three system configurations: Base SLM, Frontier, and ConvFill (except for MCP which had only Frontier and ConvFill). The presentation order of these configurations was also counterbalanced across participants to control for ordering effects. \talker and \reasoner models were drawn from two pools (Llama 3.2, Qwen3 0.6B, and Gemma 270M for \talker; Claude Opus 4.7 and OpenAI GPT-5.5 for \reasoner) and assigned to participants in a rotating manner. Each participant encountered all three conditions and all three model types (MCP for Base SLM was skipped), with the specific ordering of both conditions and models varied to prevent position bias from confounding condition or model preference judgments.

\subsection{Live Interaction Setup} 
\label{appx:subsec:live-interaction-setup}

During the study, participants interacted with a web-based voice system. The system features a press-and-hold button through which the participant records audio to send to the system. The speech-to-text service that we use is faster-whisper~\cite{fasterwhisper}, a re-implementation of OpenAI's Whisper built on the CTranslate2 inference engine. We deploy the base Whisper model ($\sim$74M parameters) running on CPU with INT8 quantization and a beam size of 1 (greedy decoding). Voice activity detection is enabled via the Silero VAD integration~\cite{silero_vad} to filter non-speech regions before transcription. 

Spoken responses are synthesized on the host machine via the macOS \texttt{say} command, which delegates to Apple's system Speech Synthesis framework~\cite{noauthor_speech_nodate}. We used the host's default system voice at the voice's default speaking rate. LLM output is stripped of Markdown before being passed to \texttt{say}.

\subsubsection{Live Interaction Example}
\label{appx:subsubsec:live-interaction-example}
The following shows a user view of the system during interaction. All Base SLM models and ConvFill \talker{} models are run with INT8 quantization through MLX on an Apple M2 SoC. Although there was an option to type an input, all participants used voice input to simulate realistic voice-based interaction.
\begin{figure*}[h]
    \centering
    \includegraphics[scale=0.5]{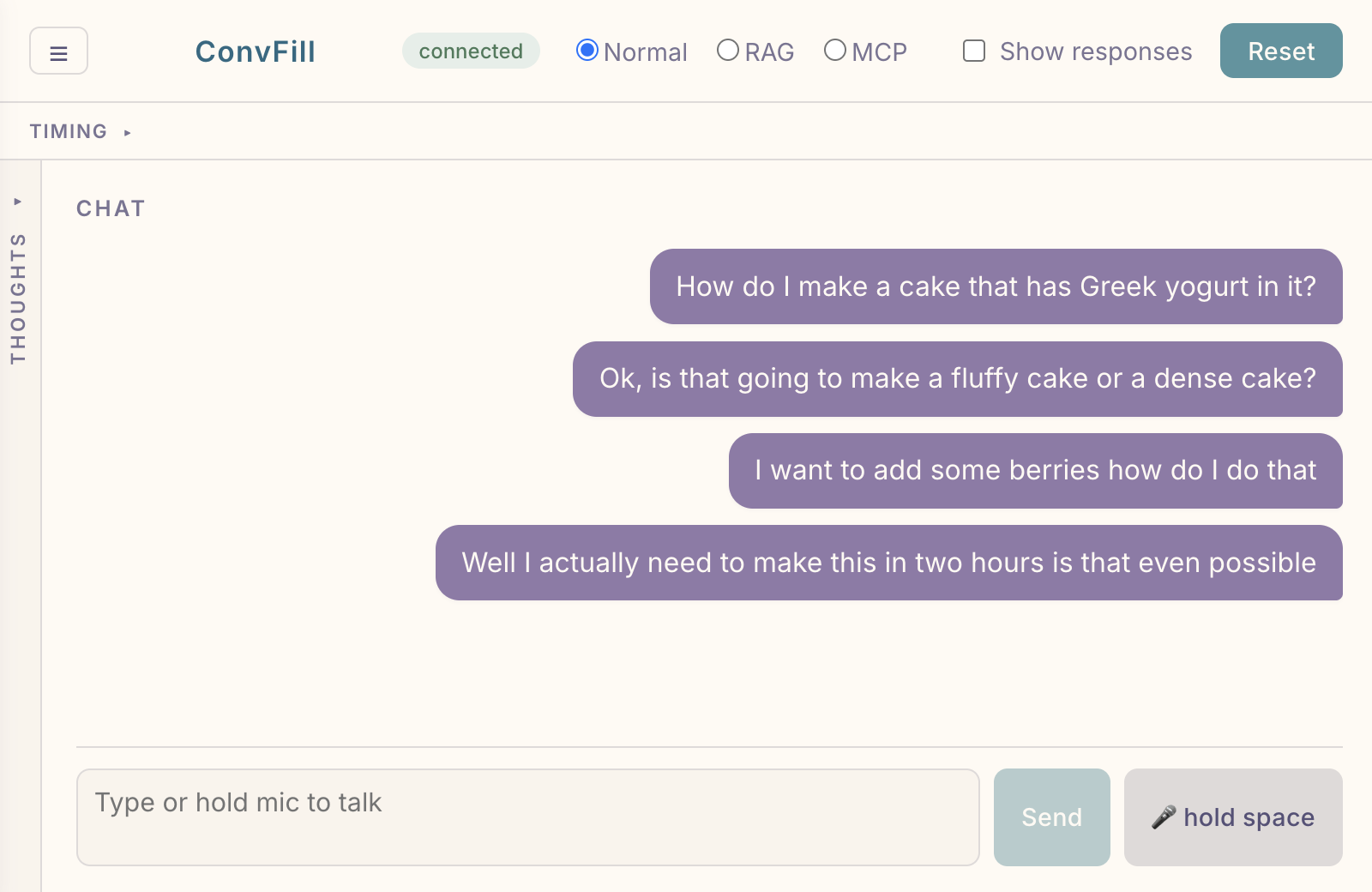} 
    \caption{\textbf{Live interaction system.} Participants see their transcribed text but do not see the model responses; they are only able to hear the responses spoken out loud.} 
    \label{fig:live-interaction-example}
\end{figure*}

\subsection{Infill Grading}
\label{appx:subsec:infill-grading}
After their live interactions in the study, participants were each asked to grade 12 samples consisting of a user utterance, a set of external knowledge chunks, and a set of conversational responses. Below is an example of a sample that was provided to the participants.\\

\noindent\textbf{User:} Try anything with Indian food; that would work as well. I will need the address if you find one, thanks!
 
\vspace{1em}

\begin{table}[h]
    \centering
\begin{tabular}{|p{0.45\textwidth}|p{0.45\textwidth}|}
\hline
\textbf{Thoughts} & \textbf{Conversational Response} \\
\hline
\vspace{0.2em}
\begin{itemize}
  \setlength\itemsep{0pt}
  \setlength\parskip{0pt}
  \item I have found 4 Indian restaurants on the east side.
  \item There are two moderate and two expensive.
  \item Did you have a price range in mind?
\end{itemize}
&
\vspace{0.2em}
Indian food, great choice! Let me look into what's available in the area. I found four Indian restaurants on the east side. There are two moderate and two expensive options. Did you have a price range in mind? \\
\hline
\end{tabular}
\caption{\textbf{Example of infill grading.} Samples of this format were provided to participants for grading.}
\label{appx:tab:infill-grading}
\end{table}

Participants graded a total of 108 unique samples. Each sample was graded by two participants, yielding a total of 216 grading examples collected from participants. 

Equal numbers of samples were selected from each of the datasets described in this paper (27 per dataset): MultiWOZ, Everyday Conversations, LLAMA1, and SimpleQA. Each example was constructed through the following method. First, a dataset was fixed. In the case of single-turn datasets (LLAMA1 and SimpleQA), a single turn was selected without replacement. Then, a Frontier model was selected (one of Claude Opus 4.7, Gemini 3.1 Pro, and GPT-5.5), and a \talker model was randomly selected from one of the \talker models in \autoref{subsec:eval-models} without replacement, such that each model was selected at least once, to produce the conversational response conditioned on the external knowledge (referred to as ``Thoughts'' in \autoref{appx:tab:infill-grading}).

Participants were given the following descriptions of each of the metrics for grading:

\begin{itemize}[itemsep=0pt]
    \item \textbf{Coverage:} Key facts represented in the thought are reflected in the response. Given the context of the user's query, the important information from the thought is reflected in the response.
    \begin{itemize}[itemsep=0pt]
        \item[5:] All important information is in the response.
        \item[4:] Some information is missing, but it does not affect the takeaways in a major way.
        \item[3:] Some information is missing, but the user will be led in the right direction even if they do not have the complete picture.
        \item[2:] Some information is conveyed, but not the important parts.
        \item[1:] Information from the thought does not appear in the response.
    \end{itemize}
    \item \textbf{Faithfulness:} Statements in the response do not contradict the thoughts or introduce additional unsupported claims.
    \begin{itemize}[itemsep=0pt]
        \item[5:] Nothing in the response contradicts the thoughts or introduces unsupported claims; there are unambiguously no distortions of meaning.
        \item[4:] Small differences arise from paraphrasing or conversational rephrasing, but they are at worst inconsequential technicalities.
        \item[3:] There is at most one minor inaccuracy, distortion, or unsupported claim, but this would not significantly affect the user's overall understanding.
        \item[2:] There is a clear inaccuracy, fabrication, or distortion that would lead the listener to a wrong conclusion on some part of the query.
        \item[1:] There are multiple contradictions or fabrications; the listener leaves with a substantively wrong understanding.
    \end{itemize}
    \item \textbf{Helpfulness:} The conversational response answers the user's query. This relates to the form and structure of the response, not whether it appropriately incorporates the contents of the thoughts.
    \begin{itemize}[itemsep=0pt]
        \item[Yes:] The response answers the user's question or statement, regardless of whether it is correct or incorporates the thoughts.
        \item[No:] The response does not answer the user's question, even if it accurately incorporates the thoughts.
    \end{itemize}
\end{itemize}

Inter-rater reliability was strong across all three evaluation dimensions (Coverage (1--5),
Faithfulness (1--5),
Helpfulness (Y/N)), with Krippendorff's $\alpha$ of 0.672 (ordinal) for Coverage, 0.67 (ordinal) for Faithfulness, and 0.751 (nominal) for Helpfulness, indicating substantial agreement among human annotators.

We also calculated human-LLM agreement with the grading presented in \autoref{appx:subsec:infill-grading} for Coverage, Faithfulness, and Helpfulness. Human-LLM agreement was moderate, with Kendall's $\tau_b$ of 0.55 and 0.554 for the ordinal Coverage and Faithfulness dimensions, respectively, and point-biserial $r$ of 0.69 for the binary Helpfulness dimension, with threshold agreement reaching 91.70\% for Helpfulness at a mean absolute distance of 0.332--0.494 across ordinal scales.
\newpage
\subsection{TOST Test Details for Likert-Scale Ratings} 
\label{appx:subsec:tost-details}

\begin{table}[h]
\centering

\begin{tabular}{lcccccccc}
\toprule
\textbf{Metric} & \shortstack{\textbf{Frontier}\\\textbf{mean}} & \shortstack{\textbf{ConvFill}\\\textbf{mean}} & \textbf{Diff} & \textbf{U\textsubscript{lower}} & \textbf{p\textsubscript{lower}} & \textbf{U\textsubscript{upper}} & \textbf{p\textsubscript{upper}} & \textbf{Equiv.} \\
\midrule
Latency           & 3.46 & 4.24 & $-0.78$ & 1326.0 & .795 & 576.0  & $<$.001 & No  \\
Clarity           & 4.81 & 4.69 & $+0.13$ & 2566.0 & $<$.001 & 582.0  & $<$.001 & Yes \\
Fluency           & 4.83 & 4.56 & $+0.28$ & 2615.0 & $<$.001 & 864.0  & $<$.001 & Yes \\
Response Length   & 4.07 & 3.94 & $+0.13$ & 2000.0 & $<$.001 & 1062.0 & .007    & Yes \\
Coherence & 4.61 & 4.15 & $+0.46$ & 2385.0 & $<$.001 & 1089.0 & .010    & Yes \\
Task Completion   & 4.35 & 4.31 & $+0.04$ & 2097.0 & $<$.001 & 940.0  & $<$.001 & Yes \\
Naturalness  & 4.41 & 3.83 & $+0.57$ & 2352.0 & $<$.001 & 1441.0 & .459    & No  \\
Satisfaction      & 3.80 & 3.81 & $-0.02$ & 1863.0 & .006    & 1030.0 & .004    & Yes \\
\bottomrule
\end{tabular}
\caption{\textbf{Full TOST results.} Wilcoxon rank-sum results for Frontier vs.\ ConvFill, aggregated across all tasks ($N = 54$ per system, $\delta = 0.5$, $\alpha = .05$). Diff = Frontier mean $-$ ConvFill mean. $p_{\text{lower}}$ tests $H_0$: Frontier $-$ ConvFill $\leq -\delta$; $p_{\text{upper}}$ tests $H_0$: Frontier $-$ ConvFill $\geq +\delta$. Equivalence is declared when both $p < .05$.}
\label{tab:tost-full}
\end{table}

\subsection{Test Details for User Rankings across Tasks}
\label{appx:subsec:test-rankings}
Pairwise ranking preferences between the Frontier and ConvFill configurations were assessed using Fisher's exact tests on rank-1 counts per task (i.e., how many participants ranked each system first). For the Direct task, the Frontier configuration received 10 rank-1 votes and ConvFill received 8 ($p = .740$). For the RAG task, ConvFill received 12 rank-1 votes compared to 5 for the Frontier configuration ($p = .044$), indicating a significant preference for ConvFill. For the MCP task, the Base SLM configuration was excluded because it could not complete tool-call interactions. Of the remaining two systems, ConvFill received 11 rank-1 votes and the Frontier configuration received 7. As neither system received any rank-3 votes in the MCP task, a two-sided binomial test was used in place of Fisher's exact test, treating each participant's rank-1 assignment as a binary choice ($p = .481$).

\clearpage
\subsection{ConvFill Speedup, TTFR, \& Filler Details}
\label{appx:subsec:convfill-speedup}

This section provides a full breakdown of the conversational infill \talker{} speedups over the conversational infill \reasoner{} as well as TTFR.

\begin{table}[h]
\centering
\begin{tabular}{lcccc}
\toprule
\textbf{Task} & \textbf{n} & \textbf{Mean ($\times$)} & \textbf{SD ($\times$)} & \textbf{95\% CI\,$\pm$} \\
\midrule
Direct      & 82  & $7.38$  & $4.53$  & $0.99$ \\
RAG         & 73  & $9.20$  & $5.75$  & $1.34$ \\
MCP         & 93  & $19.12$ & $12.93$ & $2.66$ \\
\bottomrule
\end{tabular}
\caption{\textbf{ConvFill speedup by task.} Speedup = convfill\_reasoner\_latency / convfill\_talker\_latency. Values $>1$ indicate that the filler arrived sooner than the \reasoner{} response would have.}
\label{appx:tab:speedup}
\end{table}

\begin{table}[h]
\centering
\small
\begin{tabular}{llcc}
\toprule
\textbf{Task} & \textbf{Group} & \textbf{Mean $\pm$ SD (ms)} & \textbf{95\% CI\,$\pm$} \\
\midrule
\multirow{4}{*}{Direct}
  & Base SLM         & $617 \pm 689$         & $160$ \\
  & \talker         & $542 \pm 636$         & $140$ \\
  & \reasoner          & $2947 \pm 1459$ & $321$ \\
  & Frontier   & $2578 \pm 1321$ & $279$ \\
\midrule
\multirow{4}{*}{RAG}
  & Base SLM         & $3812 \pm 2392$ & $547$ \\
  & \talker         & $976 \pm 1177$     & $275$ \\
  & \reasoner          & $4852 \pm 1751$ & $409$ \\
  & Frontier   & $4048 \pm 1706$ & $409$ \\
\midrule
\multirow{4}{*}{MCP}
  & Base SLM         & -- & -- \\
  & \talker         & $478 \pm 300$         & $61$  \\
  & \reasoner          & $7242 \pm 3850$ & $793$ \\
  & Frontier   & $8098 \pm 4213$ & $912$ \\
\bottomrule
\end{tabular}
\caption{\textbf{Latency (ms) by task and mode.} Base SLM has no MCP condition because the base SLM cannot reliably execute tool calls. ConvFill \talker{} latency is experienced by users; ConvFill \reasoner{} latency is the model's latency.}
\label{tab:latency}
\end{table}

\begin{table}[h]
\centering
\small
\begin{tabular}{lccc}
\toprule
\textbf{Task} & \textbf{Mean \standout{<sil>}} & \textbf{Std \standout{<sil>}} & \textbf{95\% CI\,$\pm$} \\
\midrule
Direct & $1.16$ & $0.40$ & $0.09$ \\
RAG    & $1.34$ & $0.53$ & $0.12$ \\
MCP    & $2.35$ & $1.12$ & $0.23$ \\
\bottomrule
\end{tabular}
\caption{\textbf{\standout{<sil>} filler tokens per turn by task.} \standout{<sil>} token counts were recorded during the user study. Mean and standard deviation of \standout{<sil>} tokens generated per turn, with 95\% confidence interval half-widths.}
\label{tab:sil}
\end{table}

\section{Full Benchmark Evaluation Results}
\label{appx:full-qa-results}
\begin{table*}[t]
\centering
\tiny
\setlength{\tabcolsep}{1.2pt}
\begin{tabular*}{\textwidth}{@{\extracolsep{\fill}}l l r r r r r r r@{}}
\toprule
Dataset & Model / baseline & \shortstack[c]{Base SLM} & \multicolumn{2}{c}{\shortstack[c]{Claude Opus 4.7\\\talker}} & \multicolumn{2}{c}{GPT-5.5 \talker} & \multicolumn{2}{c}{\shortstack[c]{Gemini 3.1\\Pro \talker}} \\
\cmidrule(lr){4-5}\cmidrule(lr){6-7}\cmidrule(lr){8-9}
 &  &  & Acc. & \shortstack[c]{Cond.\ acc.} & Acc. & \shortstack[c]{Cond.\ acc.} & Acc. & \shortstack[c]{Cond.\ acc.} \\
\midrule
\multirow[t]{9}{*}{SimpleQA} & Frontier & -- & 44.9 [41.8,48.0] & -- & 62.7 [59.7,65.6] & -- & 73.1 [70.3,75.8] & -- \\
 & \reasoner & -- & 42.9 [39.9,46.0] & -- & 61.3 [58.2,64.3] & -- & 65.7 [62.7,68.6] & -- \\
\cmidrule(lr){2-9}
 & SmolLM2 135M & 0.4 [0.2,1.0] & 38.0 [35.0,41.0] & 87.4 [83.9,90.2] & 55.2 [52.1,58.3] & 89.1 [86.4,91.3] & 60.7 [57.6,63.7] & 91.9 [89.6,93.8] \\
 & Gemma 3 270M & 0.9 [0.5,1.7] & 40.6 [37.6,43.7] & 93.0 [90.2,95.1] & 56.6 [53.5,59.6] & 91.2 [88.7,93.2] & 61.9 [58.8,64.9] & 93.6 [91.5,95.2] \\
 & SmolLM2 360M & 1.3 [0.8,2.2] & 41.2 [38.2,44.3] & 94.2 [91.5,96.0] & 57.7 [54.6,60.7] & 93.3 [91.1,95.0] & 62.8 [59.8,65.7] & 94.8 [92.9,96.3] \\
 & Qwen3 0.6B & 1.4 [0.8,2.3] & 42.4 [39.4,45.5] & 97.2 [95.2,98.4] & 60.3 [57.2,63.3] & 97.6 [96.0,98.5] & 64.8 [61.8,67.7] & 98.5 [97.2,99.2] \\
 & Gemma 3 1B & 2.2 [1.5,3.3] & 41.5 [38.5,44.6] & 96.0 [93.7,97.5] & 59.5 [56.4,62.5] & 96.2 [94.4,97.5] & 64.3 [61.3,67.2] & 97.6 [96.1,98.5] \\
 & Llama 3.2 1B & 0.5 [0.2,1.2] & 41.3 [38.3,44.4] & 94.9 [92.4,96.6] & 57.7 [54.6,60.7] & 93.5 [91.2,95.2] & 63.7 [60.7,66.6] & 96.5 [94.8,97.7] \\
 & SmolLM2 1.7B & 2.5 [1.7,3.7] & 42.5 [39.5,45.6] & 97.2 [95.2,98.4] & 61.0 [57.9,64.0] & 98.7 [97.4,99.3] & 65.1 [62.1,68.0] & 98.6 [97.4,99.3] \\
\midrule
\multirow[t]{9}{*}{LLAMA1} & Frontier & -- & 88.0 [83.8,91.2] & -- & 85.0 [80.5,88.6] & -- & 86.3 [82.0,89.8] & -- \\
 & \reasoner & -- & 85.7 [81.2,89.2] & -- & 86.3 [82.0,89.8] & -- & 87.0 [82.7,90.3] & -- \\
\cmidrule(lr){2-9}
 & SmolLM2 135M & 34.7 [29.5,40.2] & 83.7 [79.1,87.4] & 95.7 [92.5,97.6] & 82.7 [78.0,86.5] & 95.4 [92.1,97.3] & 85.7 [81.2,89.2] & 97.7 [95.1,98.9] \\
 & Gemma 3 270M & 40.7 [35.3,46.3] & 80.7 [75.8,84.7] & 91.4 [87.4,94.3] & 81.3 [76.5,85.3] & 93.1 [89.3,95.6] & 81.7 [76.9,85.6] & 92.7 [88.9,95.3] \\
 & SmolLM2 360M & 51.3 [45.7,56.9] & 84.3 [79.8,88.0] & 97.7 [95.0,98.9] & 85.0 [80.5,88.6] & 97.7 [95.0,98.9] & 85.3 [80.9,88.9] & 97.7 [95.1,98.9] \\
 & Qwen3 0.6B & 41.3 [35.9,47.0] & 84.7 [80.2,88.3] & 96.9 [94.0,98.4] & 84.0 [79.4,87.7] & 96.1 [93.0,97.9] & 85.7 [81.2,89.2] & 97.3 [94.6,98.7] \\
 & Gemma 3 1B & 63.0 [57.4,68.3] & 85.7 [81.2,89.2] & 97.3 [94.5,98.7] & 84.7 [80.2,88.3] & 96.9 [94.0,98.4] & 84.7 [80.2,88.3] & 96.2 [93.1,97.9] \\
 & Llama 3.2 1B & 72.0 [66.7,76.8] & 84.7 [80.2,88.3] & 96.9 [94.0,98.4] & 83.3 [78.7,87.1] & 95.4 [92.1,97.3] & 84.0 [79.4,87.7] & 95.8 [92.6,97.6] \\
 & SmolLM2 1.7B & 72.0 [66.7,76.8] & 85.0 [80.5,88.6] & 97.7 [95.0,98.9] & 86.0 [81.6,89.5] & 98.5 [96.1,99.4] & 84.7 [80.2,88.3] & 96.2 [93.1,97.9] \\
\bottomrule
\end{tabular*}
\caption{QA accuracy (\%) on SimpleQA ($N=1000$) and LLAMA1 ($N=300$). Cells report accuracy with full 95\% Wilson score confidence interval $[\mathrm{lo},\mathrm{hi}]$ over questions. For each ConvFill \talker{} column, Conditional Accuracy restricts accuracy to questions where the corresponding \reasoner{} bullet output was judged correct. Frontier and \reasoner{} rows score hosted frontier model answers and \reasoner{} outputs, respectively; the Base SLM column compares direct Base SLM responses with ConvFill responses conditioned on hosted \reasoner{} thoughts.}
\label{tab:qa_accuracy}
\end{table*}

\begin{table*}[t]
\centering
\tiny
\setlength{\tabcolsep}{1.8pt}
\begin{tabular*}{\textwidth}{@{\extracolsep{\fill}}ll l r r r r r r@{}}
\toprule
Dataset & \reasoner & Model & Entailment & Non-Contradiction & Coverage & Faithfulness & Helpfulness & Raw Helpfulness \\
\midrule
SimpleQA & \shortstack[l]{Claude Opus 4.7} & SmolLM2 135M & 90.4 [88.8,91.9] & 90.0 [88.5,91.8] & 4.91 [4.87,4.94] & 4.43 [4.36,4.49] & 0.91 [0.90,0.93] & 4.35 [4.29,4.41] \\
 &  & Gemma 3 270M & 90.3 [88.6,91.9] & 97.3 [96.5,98.1] & 4.92 [4.89,4.95] & 4.59 [4.53,4.64] & 0.96 [0.95,0.97] & 4.55 [4.50,4.60] \\
 &  & SmolLM2 360M & 95.7 [94.6,96.9] & 94.3 [93.2,95.4] & 4.97 [4.95,4.99] & 4.72 [4.67,4.77] & 0.96 [0.95,0.97] & 4.66 [4.60,4.70] \\
 &  & Qwen3 0.6B & 97.4 [96.6,98.1] & 93.6 [92.4,94.9] & 4.99 [4.98,5.00] & 4.87 [4.84,4.90] & 0.99 [0.98,0.99] & 4.83 [4.80,4.86] \\
 &  & Gemma 3 1B & 95.8 [94.8,97.0] & 92.1 [90.6,93.5] & 4.97 [4.95,4.99] & 4.76 [4.72,4.80] & 0.98 [0.97,0.99] & 4.73 [4.69,4.77] \\
 &  & Llama 3.2 1B & 97.1 [96.2,98.0] & 89.6 [88.2,91.2] & 4.99 [4.97,5.00] & 4.74 [4.70,4.78] & 0.98 [0.97,0.99] & 4.75 [4.71,4.79] \\
 &  & SmolLM2 1.7B & 97.2 [96.4,98.0] & 95.8 [94.8,96.8] & 4.99 [4.97,5.00] & 4.89 [4.86,4.92] & 0.99 [0.99,1.00] & 4.84 [4.81,4.87] \\
\cmidrule(lr){2-9}
 & GPT-5.5 & SmolLM2 135M & 90.3 [88.5,92.0] & 89.7 [87.9,91.5] & 4.89 [4.85,4.92] & 4.46 [4.39,4.53] & 0.91 [0.89,0.93] & 4.36 [4.30,4.42] \\
 &  & Gemma 3 270M & 88.0 [86.0,89.8] & 98.3 [97.5,99.0] & 4.85 [4.81,4.90] & 4.54 [4.49,4.60] & 0.92 [0.90,0.94] & 4.45 [4.39,4.51] \\
 &  & SmolLM2 360M & 95.5 [94.2,96.7] & 95.5 [94.4,96.5] & 4.95 [4.92,4.97] & 4.72 [4.66,4.77] & 0.95 [0.94,0.96] & 4.64 [4.59,4.69] \\
 &  & Qwen3 0.6B & 96.5 [95.5,97.5] & 94.0 [92.6,95.3] & 4.98 [4.96,4.99] & 4.86 [4.82,4.89] & 0.98 [0.97,0.99] & 4.82 [4.78,4.85] \\
 &  & Gemma 3 1B & 95.2 [94.0,96.5] & 92.6 [91.1,94.1] & 4.97 [4.95,4.99] & 4.75 [4.71,4.79] & 0.97 [0.96,0.98] & 4.74 [4.70,4.78] \\
 &  & Llama 3.2 1B & 94.2 [92.8,95.5] & 89.0 [87.2,90.7] & 4.98 [4.96,4.99] & 4.65 [4.60,4.70] & 0.96 [0.95,0.97] & 4.68 [4.63,4.73] \\
 &  & SmolLM2 1.7B & 98.3 [97.5,99.0] & 96.9 [95.9,97.8] & 4.98 [4.96,4.99] & 4.92 [4.90,4.95] & 0.99 [0.98,0.99] & 4.88 [4.85,4.91] \\
\cmidrule(lr){2-9}
 & \shortstack[l]{Gemini 3.1 Pro} & SmolLM2 135M & 89.4 [87.8,90.9] & 88.9 [87.3,90.4] & 4.74 [4.68,4.80] & 4.36 [4.28,4.43] & 0.88 [0.86,0.90] & 4.24 [4.17,4.31] \\
 &  & Gemma 3 270M & 90.9 [89.6,92.3] & 96.0 [95.0,96.9] & 4.74 [4.68,4.80] & 4.50 [4.44,4.56] & 0.91 [0.90,0.93] & 4.39 [4.32,4.46] \\
 &  & SmolLM2 360M & 94.9 [93.9,95.9] & 93.3 [92.2,94.5] & 4.80 [4.74,4.85] & 4.60 [4.54,4.65] & 0.91 [0.90,0.93] & 4.48 [4.41,4.55] \\
 &  & Qwen3 0.6B & 95.8 [94.8,96.6] & 93.5 [92.2,94.7] & 4.80 [4.75,4.85] & 4.75 [4.71,4.79] & 0.94 [0.93,0.95] & 4.63 [4.57,4.68] \\
 &  & Gemma 3 1B & 95.9 [94.9,96.8] & 92.0 [90.6,93.3] & 4.80 [4.74,4.85] & 4.67 [4.61,4.72] & 0.94 [0.92,0.95] & 4.55 [4.49,4.62] \\
 &  & Llama 3.2 1B & 95.1 [94.0,96.0] & 89.1 [87.5,90.7] & 4.79 [4.74,4.84] & 4.58 [4.52,4.63] & 0.93 [0.91,0.94] & 4.54 [4.48,4.60] \\
 &  & SmolLM2 1.7B & 95.5 [94.5,96.4] & 94.6 [93.6,95.6] & 4.80 [4.75,4.86] & 4.81 [4.77,4.85] & 0.95 [0.93,0.96] & 4.67 [4.61,4.73] \\
\midrule
LLAMA1 & \shortstack[l]{Claude Opus 4.7} & SmolLM2 135M & 95.8 [93.9,97.3] & 94.8 [93.1,96.4] & 4.99 [4.97,5.00] & 4.63 [4.53,4.71] & 0.98 [0.97,0.99] & 4.61 [4.53,4.68] \\
 &  & Gemma 3 270M & 94.3 [92.7,95.8] & 97.3 [95.8,98.6] & 4.97 [4.93,4.99] & 4.59 [4.48,4.68] & 0.95 [0.92,0.97] & 4.63 [4.54,4.72] \\
 &  & SmolLM2 360M & 97.3 [96.1,98.3] & 95.4 [93.1,97.4] & 5.00 [5.00,5.00] & 4.86 [4.79,4.91] & 0.99 [0.98,1.00] & 4.83 [4.77,4.88] \\
 &  & Qwen3 0.6B & 95.8 [94.4,97.2] & 90.6 [87.2,93.8] & 5.00 [5.00,5.00] & 4.83 [4.76,4.89] & 0.98 [0.96,0.99] & 4.85 [4.79,4.91] \\
 &  & Gemma 3 1B & 96.4 [95.1,97.8] & 87.1 [83.4,90.6] & 5.00 [5.00,5.00] & 4.84 [4.77,4.89] & 0.99 [0.98,1.00] & 4.87 [4.82,4.92] \\
 &  & Llama 3.2 1B & 96.3 [94.9,97.5] & 89.8 [86.8,92.9] & 5.00 [5.00,5.00] & 4.78 [4.72,4.84] & 0.99 [0.98,1.00] & 4.87 [4.82,4.92] \\
 &  & SmolLM2 1.7B & 96.4 [94.8,97.7] & 98.6 [97.8,99.4] & 5.00 [5.00,5.00] & 4.88 [4.83,4.93] & 0.99 [0.98,1.00] & 4.91 [4.86,4.94] \\
\cmidrule(lr){2-9}
 & GPT-5.5 & SmolLM2 135M & 92.9 [90.2,95.4] & 93.1 [90.7,95.3] & 4.92 [4.85,4.98] & 4.67 [4.59,4.75] & 0.96 [0.94,0.98] & 4.48 [4.39,4.57] \\
 &  & Gemma 3 270M & 90.7 [87.6,93.6] & 97.1 [95.3,98.6] & 4.93 [4.87,4.98] & 4.72 [4.63,4.80] & 0.96 [0.93,0.98] & 4.61 [4.52,4.70] \\
 &  & SmolLM2 360M & 95.1 [92.9,97.3] & 94.0 [91.1,96.5] & 5.00 [5.00,5.00] & 4.83 [4.76,4.89] & 0.99 [0.97,1.00] & 4.75 [4.69,4.81] \\
 &  & Qwen3 0.6B & 94.7 [92.1,96.9] & 89.1 [85.3,92.6] & 5.00 [5.00,5.00] & 4.88 [4.82,4.93] & 0.99 [0.98,1.00] & 4.86 [4.80,4.92] \\
 &  & Gemma 3 1B & 96.4 [94.4,98.1] & 85.6 [81.5,89.6] & 5.00 [4.99,5.00] & 4.87 [4.81,4.92] & 0.99 [0.98,1.00] & 4.85 [4.80,4.90] \\
 &  & Llama 3.2 1B & 96.2 [94.3,97.9] & 87.8 [84.3,91.4] & 5.00 [5.00,5.00] & 4.76 [4.69,4.83] & 0.99 [0.98,1.00] & 4.81 [4.75,4.87] \\
 &  & SmolLM2 1.7B & 95.2 [93.0,97.2] & 97.9 [96.2,99.2] & 5.00 [5.00,5.00] & 4.88 [4.82,4.93] & 0.99 [0.98,1.00] & 4.87 [4.82,4.91] \\
\cmidrule(lr){2-9}
 & \shortstack[l]{Gemini 3.1 Pro} & SmolLM2 135M & 96.5 [95.0,98.0] & 94.5 [92.7,96.1] & 4.99 [4.98,5.00] & 4.75 [4.67,4.82] & 0.97 [0.96,0.99] & 4.67 [4.59,4.74] \\
 &  & Gemma 3 270M & 96.3 [94.7,97.8] & 95.8 [93.8,97.5] & 4.99 [4.97,5.00] & 4.73 [4.64,4.80] & 0.96 [0.94,0.98] & 4.73 [4.66,4.80] \\
 &  & SmolLM2 360M & 94.6 [92.8,96.2] & 95.0 [92.5,97.2] & 5.00 [5.00,5.00] & 4.85 [4.78,4.90] & 0.99 [0.98,1.00] & 4.78 [4.72,4.84] \\
 &  & Qwen3 0.6B & 95.8 [93.9,97.6] & 90.8 [87.7,93.8] & 5.00 [5.00,5.00] & 4.90 [4.84,4.95] & 0.99 [0.97,1.00] & 4.87 [4.82,4.92] \\
 &  & Gemma 3 1B & 96.1 [94.4,97.7] & 87.7 [84.4,91.1] & 5.00 [5.00,5.00] & 4.88 [4.82,4.93] & 0.99 [0.98,1.00] & 4.88 [4.83,4.92] \\
 &  & Llama 3.2 1B & 97.7 [96.2,98.9] & 90.7 [87.9,93.4] & 4.99 [4.98,5.00] & 4.84 [4.78,4.89] & 0.99 [0.99,1.00] & 4.87 [4.83,4.92] \\
 &  & SmolLM2 1.7B & 95.9 [94.3,97.4] & 97.9 [96.4,99.0] & 5.00 [5.00,5.00] & 4.87 [4.81,4.93] & 0.99 [0.98,1.00] & 4.88 [4.83,4.93] \\
\bottomrule
\end{tabular*}
\caption{Thought-conditioned ConvFill \talker quality on QA datasets using live hosted \reasoner{} thoughts. Entailment is the NLI entailment rate for response fragments against their paired thoughts. Cells report mean with full 95\% bootstrap confidence interval $[\mathrm{lo},\mathrm{hi}]$ over questions; NLI values are percentage points and judge values are on the displayed scales. Bootstrap intervals resample the question unit consistently for fractional NLI scores, Likert means, and thresholded helpfulness. Raw Helpfulness is included for transparency but not used in analyses.}
\label{tab:qa_accuracy_metrics}
\end{table*}

\begin{table*}[t]
\centering
\scriptsize
\setlength{\tabcolsep}{2.2pt}
\begin{tabular*}{\textwidth}{@{\extracolsep{\fill}}l l r r r r r r@{}}
\toprule
Dataset & Model & Entailment & Non-Contradiction & Coverage & Faithfulness & Helpfulness & Raw Helpfulness \\
\midrule
MultiWOZ & SmolLM2 135M & 83.9 [82.0,85.9] & 85.9 [83.9,87.6] & 4.83 [4.79,4.87] & 4.57 [4.51,4.61] & 0.91 [0.89,0.93] & 4.23 [4.16,4.29] \\
 & Gemma 3 270M & 82.9 [80.9,84.7] & 81.0 [79.1,83.0] & 4.85 [4.81,4.89] & 4.55 [4.49,4.61] & 0.92 [0.91,0.94] & 4.27 [4.21,4.33] \\
 & SmolLM2 360M & 82.8 [80.9,84.6] & 84.5 [82.6,86.4] & 4.86 [4.82,4.90] & 4.68 [4.63,4.72] & 0.94 [0.92,0.95] & 4.34 [4.28,4.40] \\
 & Qwen3 0.6B & 83.0 [81.2,84.8] & 81.8 [79.7,83.6] & 4.90 [4.87,4.93] & 4.74 [4.69,4.78] & 0.94 [0.93,0.96] & 4.43 [4.37,4.48] \\
 & Gemma 3 1B & 82.3 [80.5,84.2] & 81.2 [79.2,83.2] & 4.85 [4.81,4.89] & 4.52 [4.46,4.58] & 0.92 [0.90,0.93] & 4.26 [4.20,4.32] \\
 & Llama 3.2 1B & 84.5 [82.8,86.3] & 81.3 [79.1,83.2] & 4.87 [4.83,4.90] & 4.60 [4.55,4.65] & 0.93 [0.91,0.95] & 4.33 [4.28,4.40] \\
 & SmolLM2 1.7B & 83.6 [81.9,85.4] & 82.8 [80.9,84.7] & 4.89 [4.86,4.93] & 4.74 [4.70,4.78] & 0.95 [0.94,0.96] & 4.46 [4.41,4.51] \\
\midrule
\shortstack[l]{Everyday\\Conversations} & SmolLM2 135M & 92.3 [88.7,95.8] & 95.6 [93.1,97.9] & 4.99 [4.97,5.00] & 4.87 [4.79,4.94] & 1.00 [1.00,1.00] & 4.83 [4.76,4.90] \\
 & Gemma 3 270M & 85.1 [79.6,89.8] & 96.5 [93.8,98.6] & 4.99 [4.98,5.00] & 4.77 [4.66,4.86] & 0.99 [0.96,1.00] & 4.86 [4.77,4.94] \\
 & SmolLM2 360M & 91.4 [87.0,95.4] & 97.9 [95.8,99.5] & 5.00 [5.00,5.00] & 4.91 [4.84,4.97] & 1.00 [1.00,1.00] & 4.93 [4.87,4.97] \\
 & Qwen3 0.6B & 92.0 [88.7,95.8] & 96.7 [93.9,98.9] & 5.00 [5.00,5.00] & 4.88 [4.79,4.95] & 1.00 [1.00,1.00] & 4.91 [4.85,4.96] \\
 & Gemma 3 1B & 84.8 [79.8,90.1] & 98.2 [96.2,99.8] & 5.00 [5.00,5.00] & 4.91 [4.84,4.97] & 1.00 [1.00,1.00] & 4.96 [4.91,4.99] \\
 & Llama 3.2 1B & 87.8 [83.5,92.3] & 97.7 [95.9,99.3] & 5.00 [5.00,5.00] & 4.86 [4.77,4.93] & 1.00 [1.00,1.00] & 4.89 [4.83,4.95] \\
 & SmolLM2 1.7B & 85.1 [80.4,89.9] & 99.1 [97.8,100.0] & 5.00 [5.00,5.00] & 4.97 [4.94,4.99] & 1.00 [1.00,1.00] & 4.99 [4.97,5.00] \\
\bottomrule
\end{tabular*}
\caption{Thought-conditioned ConvFill~\talker quality on sampled multi-turn dialogue turns using dataset-derived conversation thoughts. One non-edge turn is sampled per dialogue for grading; earlier turns in the same dialogue are generated only as hidden warm-up to preserve ConvFill self-history. Metrics and setup are identical to \autoref{subsec:multi-turn-eval}, but with more detailed confidence intervals. Cells report mean with full 95\% bootstrap confidence interval $[\mathrm{lo},\mathrm{hi}]$ over sampled turns; NLI values are percentage points and judge values are on the displayed scales. Bootstrap intervals resample the sampled-turn unit consistently for fractional NLI scores, Likert means, and thresholded helpfulness. Raw Helpfulness is included for transparency but not used in analyses.}
\label{tab:multiturn_sampled_nli_llm_judge}
\end{table*}

\begin{table*}[t]
\centering
\scriptsize
\setlength{\tabcolsep}{2.2pt}
\renewcommand{\arraystretch}{1.06}
\resizebox{\textwidth}{!}{%
\begin{tabular}{@{}llllccccccc@{}}
\toprule
\textbf{Trend} & \textbf{Setting} & \textbf{Dataset} & \textbf{\reasoner} & \textbf{QA acc.} & \textbf{Cond. acc.} & \textbf{Ent.} & \textbf{Non-C.} & \textbf{Help.} & \textbf{Coverage} & \textbf{Faith.} \\
\midrule
All models & Base SLM & SimpleQA & -- & \textbf{+ BY} & -- & -- & -- & -- & -- & -- \\
All models & CF dataset & SimpleQA & Data. & \textbf{+ BY} & -- & \textbf{+ BY} & ns & \textbf{+ BY} & \textbf{+ BY} & \textbf{+ BY} \\
All models & CF live & SimpleQA & Claude Opus 4.7 & \textbf{+ BY} & \textbf{+ BY} & \textbf{+ BY} & ns & \textbf{+ BY} & \textbf{+ BY} & \textbf{+ BY} \\
All models & CF live & SimpleQA & GPT-5.5 & \textbf{+ BY} & \textbf{+ BY} & \textbf{+ BY} & ns & \textbf{+ BY} & \textbf{+ BY} & \textbf{+ BY} \\
All models & CF live & SimpleQA & Gemini 3.1 Pro & \textbf{+ BY} & \textbf{+ BY} & \textbf{+ BY} & ns & \textbf{+ BY} & \textbf{+ BY} & \textbf{+ BY} \\
\addlinespace
All models & Base SLM & LLAMA1 & -- & \textbf{+ BY} & -- & -- & -- & -- & -- & -- \\
All models & CF dataset & LLAMA1 & Data. & \textbf{+ BY} & -- & \textbf{+ BY} & ns & \textbf{+ BY} & \textbf{+ BY} & \textbf{+ BY} \\
All models & CF live & LLAMA1 & Claude Opus 4.7 & ns & ns & ns & ns & \textbf{+ BY} & ns & \textbf{+ BY} \\
All models & CF live & LLAMA1 & GPT-5.5 & ns & ns & \textbf{+ BY} & ns & \textbf{+ BY} & \textbf{+ BY} & \textbf{+ BY} \\
All models & CF live & LLAMA1 & Gemini 3.1 Pro & ns & ns & ns & ns & \textbf{+ BY} & ns & \textbf{+ BY} \\
\addlinespace
All models & CF sampled & MultiWOZ & -- & -- & -- & ns & \textbf{- BY} & \textbf{+ BY} & \textbf{+ BY} & \textbf{+ BY} \\
\addlinespace
All models & CF sampled & Everyday & -- & -- & -- & ns & ns & ns & ns & ns \\
\midrule
SmolLM & Base SLM & SimpleQA & -- & \textbf{+ BY} & -- & -- & -- & -- & -- & -- \\
SmolLM & CF dataset & SimpleQA & Data. & \textbf{+ BY} & -- & \textbf{+ BY} & \textbf{+ BY} & \textbf{+ BY} & \textbf{+ BY} & \textbf{+ BY} \\
SmolLM & CF live & SimpleQA & Claude Opus 4.7 & \textbf{+ BY} & \textbf{+ BY} & \textbf{+ BY} & \textbf{+ BY} & \textbf{+ BY} & \textbf{+ BY} & \textbf{+ BY} \\
SmolLM & CF live & SimpleQA & GPT-5.5 & \textbf{+ BY} & \textbf{+ BY} & \textbf{+ BY} & \textbf{+ BY} & \textbf{+ BY} & \textbf{+ BY} & \textbf{+ BY} \\
SmolLM & CF live & SimpleQA & Gemini 3.1 Pro & \textbf{+ BY} & \textbf{+ BY} & \textbf{+ BY} & \textbf{+ BY} & \textbf{+ BY} & \textbf{+ BY} & \textbf{+ BY} \\
\addlinespace
SmolLM & Base SLM & LLAMA1 & -- & \textbf{+ BY} & -- & -- & -- & -- & -- & -- \\
SmolLM & CF dataset & LLAMA1 & Data. & \textbf{+ BY} & -- & \textbf{+ BY} & \textbf{+ BY} & \textbf{+ BY} & \textbf{+ BY} & \textbf{+ BY} \\
SmolLM & CF live & LLAMA1 & Claude Opus 4.7 & ns & ns & ns & \textbf{+ BY} & ns & ns & \textbf{+ BY} \\
SmolLM & CF live & LLAMA1 & GPT-5.5 & ns & ns & ns & \textbf{+ BY} & ns & \textbf{+ BY} & \textbf{+ BY} \\
SmolLM & CF live & LLAMA1 & Gemini 3.1 Pro & ns & ns & ns & ns & ns & ns & ns \\
\addlinespace
SmolLM & CF sampled & MultiWOZ & -- & -- & -- & ns & \textbf{- BY} & \textbf{+ BY} & \textbf{+ BY} & \textbf{+ BY} \\
\addlinespace
SmolLM & CF sampled & Everyday & -- & -- & -- & ns & ns & ns & ns & ns \\
\midrule
Gemma & Base SLM & SimpleQA & -- & ns & -- & -- & -- & -- & -- & -- \\
Gemma & CF dataset & SimpleQA & Data. & \textbf{+ BY} & -- & \textbf{+ BY} & \textbf{- BY} & \textbf{+ BY} & \textbf{+ BY} & \textbf{+ BY} \\
Gemma & CF live & SimpleQA & Claude Opus 4.7 & ns & ns & \textbf{+ BY} & \textbf{- BY} & ns & ns & \textbf{+ BY} \\
Gemma & CF live & SimpleQA & GPT-5.5 & \textbf{+ BY} & \textbf{+ BY} & \textbf{+ BY} & \textbf{- BY} & \textbf{+ BY} & \textbf{+ BY} & \textbf{+ BY} \\
Gemma & CF live & SimpleQA & Gemini 3.1 Pro & \textbf{+ BY} & \textbf{+ BY} & \textbf{+ BY} & \textbf{- BY} & \textbf{+ BY} & \textbf{+ BY} & \textbf{+ BY} \\
\addlinespace
Gemma & Base SLM & LLAMA1 & -- & \textbf{+ BY} & -- & -- & -- & -- & -- & -- \\
Gemma & CF dataset & LLAMA1 & Data. & \textbf{+ BY} & -- & \textbf{+ BY} & \textbf{- BY} & \textbf{+ BY} & \textbf{+ BY} & \textbf{+ BY} \\
Gemma & CF live & LLAMA1 & Claude Opus 4.7 & \textbf{+ BY} & \textbf{+ BY} & ns & \textbf{- BY} & \textbf{+ BY} & ns & \textbf{+ BY} \\
Gemma & CF live & LLAMA1 & GPT-5.5 & ns & ns & \textbf{+ BY} & \textbf{- BY} & \textbf{+ BY} & ns & \textbf{+ BY} \\
Gemma & CF live & LLAMA1 & Gemini 3.1 Pro & ns & ns & ns & \textbf{- BY} & ns & ns & \textbf{+ BY} \\
\addlinespace
Gemma & CF sampled & MultiWOZ & -- & -- & -- & ns & ns & ns & ns & ns \\
\addlinespace
Gemma & CF sampled & Everyday & -- & -- & -- & ns & ns & ns & ns & ns \\
\bottomrule
\end{tabular}%
}
\caption{Scaling-test comparison using conservative Benjamini--Yekutieli FDR correction~\cite{benjamini_control_2001}. Cells report trend direction only when BY-significant; ns = not BY-significant, and -- = not applicable. \textit{All models} denotes an exploratory parameter trend across all seven \talker{} models and is shown first for orientation; SmolLM is the primary within-family trend, and Gemma is a two-point 270M-to-1B contrast. CF = ConvFill; Data. = dataset/reference thoughts; QA acc. = Accuracy; Cond. acc. = Conditional Accuracy; Ent. = Entailment; Non-C. = Non-Contradiction; Help. = Helpfulness; Faith. = Faithfulness.}
\label{tab:family_scaling_comparison}
\end{table*}

\clearpage

\section{Inference and Evaluation Prompts}
\label{appx:sec:inf-and-eval-prompts}
This section of the appendix contains the exact per-task prompts used during live interaction and the LLM-as-a-Judge prompt used for single- and multi-turn evaluation.

\subsection{LLM-as-a-Judge Prompt}
\label{appx:subsec:llm-judge-prompt}

\begin{prompt}[title={Prompt \thetcbcounter: LLM-as-a-Judge Prompt}, label=prompt:llm-as-a-judge-prompt]
\begin{lstlisting}[style=convfillprompt]
You are evaluating ONE turn of a ConvFill conversation, a real-time two-model
system. A reasoner LLM produces a stream of factual "thoughts" (bullet-like
knowledge), and a small talker model weaves those thoughts into a fluent
spoken response. Your job is to rate this **single turn** across three rubric
dimensions.

# Rubric

Single-turn evaluation. All metrics on a 1-5 Likert scale, 5 = best.

You are given exactly one user query, the bullet-list of thoughts that were
streamed in for that turn, and the conversational response the system produced. Word-for-word match is not required -- what matters is whether
the listener leaves this turn with the right understanding and can act on
it.

**Use the thoughts as the only factual source.** Do not penalize the response
for failing to answer parts of the user query that are not answered by the
thoughts. The task is to judge how well the response conveys the available
thoughts, not whether the reasoner thoughts themselves were complete.

Keep the three dimensions separate:

- Coverage is about whether the response includes the relevant thought
  content. Ignore whether that content is phrased awkwardly or attached to
  the wrong relation; score those problems under Faithfulness instead.
  The denominator is the thoughts, not every part of the user question: if
  the user asks two things but the thoughts answer only one, Coverage asks
  whether the response conveyed that one available answer.
- Faithfulness is about whether the response changes, contradicts, or adds
  unsupported meaning beyond the thoughts.
- Helpfulness is about whether a normal user would consider this response 
  aligned and structurally consistent with their query, independent of 
  the other two dimensions. 

## 1. Coverage

*Does the response convey the relevant information expressed in the thoughts?*

A fact in the thoughts is "load-bearing" if it is relevant to the user query
or is the main available answer. Score on whether that available information
is present in the response, not whether every part of the user's query was
answerable from the thoughts.
Paraphrase, reordering, compression, unit/format conversion, and casual
wording are fine.

This metric is about errors of *omission* only. If the response includes the
right value or entity but puts it in a clumsy or partly wrong frame, give
Coverage credit for including it and use Faithfulness to penalize the
distortion. Do not lower Coverage merely because the response adds filler,
extra unsupported phrasing, or conversational scaffolding.

When the thoughts themselves are incomplete, do **not** lower Coverage for
missing facts that are absent from the thoughts. If the thoughts answer only
one part of a multi-part user query, Coverage asks whether the response
conveys that available part.

\end{lstlisting}
\end{prompt}

\begin{prompt}[title={Prompt \thetcbcounter: LLM-as-a-Judge Prompt contd. (extends \autoref{prompt:llm-as-a-judge-prompt})}, label=prompt:llm-as-a-judge-prompt-2]
\begin{lstlisting}[style=convfillprompt]
If the key answer value, name, number, or entity from the thoughts appears in
the response, give substantial Coverage credit even when the response attaches
it to the wrong relation or frames it awkwardly. Relation errors belong under
Faithfulness. For short symbolic answers, treat capitalization and spacing
differences as surface form for Coverage.

- **5** -- Every relevant thought fact is conveyed in the response.
- **4** -- All relevant thought facts are conveyed; only minor,
  non-load-bearing details from the thoughts are omitted.
- **3** -- Most relevant thought facts are present, but one material available
  fact is missing.
- **2** -- Multiple relevant thought facts are missing; the response materially
  under-informs the listener relative to the available thoughts.
- **1** -- Most of the available thought content is absent from the response.

## 2. Faithfulness

*Does the response avoid contradicting or fabricating beyond the thoughts?*

This metric is about errors of *commission* -- distortions, fabrications,
unsupported claims, misleading action claims, or framings that change a fact's
meaning. A response that merely omits information can still score high here,
but a response that says it is looking up, pulling up, checking, or already
has details while giving no answer is not a clean omission; it is a misleading
non-answer. Severity is weighted by how much the distortion affects the
available answer: a fabrication on the main answer is worse than one in a
tangential aside.

Be forgiving about surface form. Do not penalize capitalization, punctuation,
spacing, or harmless formatting differences when the intended answer is clear.
For times, accept an unambiguous 12-hour rendering of a 24-hour time when the
conversation context makes the meaning clear. Do not treat missing AM/PM as a
contradiction when the surrounding user query or response makes the time of
day clear enough.
Do not penalize unfounded but non-factual connective language unless it asserts
or implies an unsupported fact.

For short names, codes, and style symbols, do not over-penalize
capitalization or spacing differences when the same letters are present and a
normal listener/reader would identify the intended answer.

Process filler such as "let me look that up," "I found the details," or "I
have it right here" is low-value conversational scaffolding. If the response
also gives the same relayed answer and the filler does not change the
factual meaning, do not penalize the filler; the Faithfulness score should
usually remain **5**. If the response is mostly such process filler and never
actually gives the available answer, Faithfulness should usually be around
**3** rather than 5: it did not contradict the thoughts, but it created a
misleading impression of progress while withholding the actual answer.

- **5** -- Nothing in the response contradicts the thoughts or introduces
  unsupported claims; unambiguously no distortions of meaning.
- **4** -- Overall faithful. Small differences arise from paraphrasing or
  conversational rephrasing, but they are at worst inconsequential technicalities.
- **3** -- There is at most one minor inaccuracy, distortion, or unsupported claim,
  but this would not significantly cause incorrect ideas. 
- **2** -- A clear inaccuracy, fabrication, or distortion that would lead the
  listener to a wrong conclusion on some part of the statement.
- **1** -- Multiple contradictions or fabrications; the listener leaves with
  a substantively wrong understanding on key parts of the response.
\end{lstlisting}
\end{prompt}

\begin{prompt}[title={Prompt \thetcbcounter: LLM-as-a-Judge Prompt contd. (extends \autoref{prompt:llm-as-a-judge-prompt-2})}, label=prompt:llm-as-a-judge-prompt-3]
\begin{lstlisting}[style=convfillprompt]
## 3. Helpfulness

*Given the user's query, is the response structured in an appropriate fashion?*

Scores **4-5** indicate generally appropriate responses to the question,
statement, or phrase. Even if wrong, it is still syntactically an appropriate
response. Scores **1-2** count as "No, not valid or appropriate response." Use
"3" for only truly borderline cases that are still on the side of acceptable
and appropriate grammatically and given the conversation context.

A response can be helpful even if it is brief or slightly awkward. A response
is not helpful when it withholds the available answer, answers the wrong
question, gives only vague setup/filler, or states the right words in a way
that does not follow and connect to what the user said.

Factual correctness is already captured by Coverage and Faithfulness. 

If the user asks "What is the day after Monday," A **5** could be answers
like "The day after Monday is Tuesday" or "The day after Monday is Wednesday"
since both are structurally appropriate answers, even though the listed day
itself isn't exactly correct. "Tuesday is a day of the week" or "Ah, Tuesday is
something everyone always says" would be a **2**, since it does not answer the 
user's question. The fact that the **answer** is contained does not matter 
for this purpose. 

- **5** -- Appropriately answers the user. It fully fits within the structure
  of a normal and appropriate conversation. The most expected type of response.
- **4** -- Answers the query but in a form that sounds possibly odd. 
- **3** -- Still structurally useful and reasonable based on what was said, but
  only barely. 
- **2** -- Vague or tangential; does not meaningfully respond to the user's query.
- **1** -- Off-topic, evasive, or actively misaligned given what the user asked.

# Output format

Return ONLY a single JSON object, no markdown, no commentary outside the JSON.
The object must have exactly this shape:

```
{
  "coverage":     {"score": <int 1-5>, "rationale": "<1-2 sentences>"},
  "faithfulness": {"score": <int 1-5>, "rationale": "<1-2 sentences>"},
  "helpfulness":  {"score": <int 1-5>, "rationale": "<1-2 sentences>"}
}
```

Each `score` is an integer from 1 to 5 (use the anchors above). Each
`rationale` is one or two sentences citing the specific behavior that
drove the score -- name a fact, bullet, or phrase wherever possible.

# Turn

### user
{{ user_turn }}

### thoughts
{% for b in thoughts_substance %}- {{ b }}
{% endfor %}
### conversational response
{{ convfill_response }}
\end{lstlisting}
\end{prompt}

\begin{prompt}[title={Prompt \thetcbcounter: Normal Task Backend Prompt}, label=prompt:normal-task-backend-prompt]
\begin{lstlisting}[style=convfillprompt]
Your job is to take the previous turns of the conversation and respond with concise sentences, each representing one thought or phrase that can be said to the user. The user's queries were spoken out loud and transcribed via a speech-to-text (STT) model. 
Your job is not to hold a conversation or show emotion or connection, just to act as an information supplier and answer whatever the user asks to move the conversation forward. 
ONLY output concise sentences, nothing else. No em dashes. No examples. No recommendations unless asked for.
You must put all numbers and symbols in words.
Anything that must be relayed verbatim MUST be put in quotes. Max 3 sentences. Ask AT MOST ONE question.

Example Conversation and Response:
User: Hey there, I want to know how to make some cafe drinks at home to save money.
You: Making drinks at home is not too hard, don't worry. What kind of drinks do you like?
User: Hmm, I like matcha and non-coffee drinks. I'm glad they are not too hard to make at home. I really like flavored matcha drinks, like vanilla or strawberry. Are those easy to make as well?
You: The difficulty depends on how much you want to make homemade. You can just buy a Torani vanilla bean syrup for the vanilla one. You can also make a homemade strawberry puree for the strawberry matcha.

OR

You: You can just buy a Torani vanilla bean syrup for the vanilla. You can make a homemade strawberry puree for the strawberry matcha. What are you looking for in terms of effort?

OR

You: What are you looking for in terms of effort? You can just buy a Torani vanilla bean syrup for the vanilla. You can also make a homemade strawberry puree for the strawberry matcha.

Here is the conversation:
{{ conversation }}

\end{lstlisting}
\end{prompt}

\begin{prompt}[title={Prompt \thetcbcounter: RAG Task Backend Prompt}, label=prompt:rag-task-backend-prompt]
\begin{lstlisting}[style=convfillprompt]
Your job is to take the previous turns of the conversation and respond with concise sentences, each representing one thought or phrase that can be said to the user. The user's queries were spoken out loud and transcribed via a speech-to-text (STT) model. 
Your job is not to hold a conversation or show emotion or connection, just to act as an information supplier and answer whatever the user asks to move the conversation forward. 
ONLY output concise sentences, nothing else. No em dashes. No examples. No recommendations unless asked for.
You must put all numbers and symbols in words for dollar amounts and large numbers.
Anything that must be relayed verbatim MUST be put in quotes. Max 3 sentences. DO NOT ASK ANY QUESTIONS.

Example Conversation and Response:
User: Hey there, I want to know how to make some cafe drinks at home to save money.
You: Making drinks at home is not too hard, don't worry. What kind of drinks do you like?
User: Hmm, I like matcha and non-coffee drinks. I'm glad they are not too hard to make at home. I really like flavored matcha drinks, like vanilla or strawberry. Are those easy to make as well?
You: The difficulty depends on how much you want to make homemade. You can just buy a Torani vanilla bean syrup for the vanilla one. You can also make a homemade strawberry puree for the strawberry matcha.

OR

You: You can just buy a Torani vanilla bean syrup for the vanilla. You can make a homemade strawberry puree for the strawberry matcha. What are you looking for in terms of effort?

OR

You: What are you looking for in terms of effort? You can just buy a Torani vanilla bean syrup for the vanilla. You can also make a homemade strawberry puree for the strawberry matcha.

Here is the context for your response. To help you respond to queries, here are some potentially relevant excerpts from the University of Washington (UW) Computer Science PhD Student Handbook:
{{ rag_context }}

Here is the conversation:
{{ conversation }}

\end{lstlisting}
\end{prompt}

\begin{prompt}[title={Prompt \thetcbcounter: MCP Task Backend Prompt}, label=prompt:mcp-task-backend-prompt]
\begin{lstlisting}[style=convfillprompt]
Your job is to take the previous turns of the conversation and respond with concise sentences, each representing one thought or phrase that can be said to the user. The user's queries were spoken out loud and transcribed via a speech-to-text (STT) model. 
Your job is not to hold a conversation or show emotion or connection, just to act as an information supplier and answer whatever the user asks to move the conversation forward. 
ONLY output concise sentences, nothing else. No em dashes. No examples. No recommendations unless asked for.
You must put all numbers and symbols in words for dollar amounts and large numbers.
Anything that must be relayed verbatim MUST be put in quotes. Max 3 sentences. Ask at most one question.

Use the provided tools when you need to look up information. Focus on likely keywords instead of long verbatim phrases for searches (e.g., "ticket" not "the concert ticket for the show tonight").

Example Conversation and Response:
User: Hey there, I want to know how to make some cafe drinks at home to save money.
You: Making drinks at home is not too hard, don't worry. What kind of drinks do you like?
User: Hmm, I like matcha and non-coffee drinks. I'm glad they are not too hard to make at home. I really like flavored matcha drinks, like vanilla or strawberry. Are those easy to make as well?
You: The difficulty depends on how much you want to make homemade. You can just buy a Torani vanilla bean syrup for the vanilla one. You can also make a homemade strawberry puree for the strawberry matcha.

OR

You: You can just buy a Torani vanilla bean syrup for the vanilla. You can make a homemade strawberry puree for the strawberry matcha. What are you looking for in terms of effort?

OR

You: What are you looking for in terms of effort? You can just buy a Torani vanilla bean syrup for the vanilla. You can also make a homemade strawberry puree for the strawberry matcha.

Here is the conversation:
{{ conversation }}

\end{lstlisting}
\end{prompt}

\end{document}